\documentclass[10pt,twocolumn,letterpaper]{article}

\usepackage[pagenumbers]{cvpr} % To force page numbers, e.g. for an arXiv version

\usepackage[dvipsnames]{xcolor}

\definecolor{cvprblue}{rgb}{0.21,0.49,0.74}
\usepackage[pagebackref,breaklinks,colorlinks,citecolor=cvprblue]{hyperref}
\usepackage{multirow}
\usepackage{stfloats}

\def\paperID{10876} % *** Enter the Paper ID here
\def\confName{CVPR}
\def\confYear{2024}

\title{Domain Generalization of 3D Object Detection by Density-Resampling}

\author{Shuangzhi Li, Lei Ma, and Xingyu Li\\
Department of Electrical and Computer Engineering, University
of Alberta, Alberta, Canada\\
{\tt\small \{shuangzh, lma7, xingyu\}@ualberta.ca}
}

\begin{document}
\maketitle

\begin{abstract}
Point-cloud-based 3D object detection suffers from performance degradation when encountering data with novel domain gaps. To tackle it, the single-domain generalization (SDG) aims to generalize the detection model trained in a limited single source domain to perform robustly on unexplored domains. In this paper, we propose an SDG method to improve the generalizability of 3D object detection to unseen target domains. 
Unlike prior SDG works for 3D object detection solely focusing on data augmentation, our work introduces a novel data augmentation method and contributes a new multi-task learning strategy in the methodology.
Specifically, from the perspective of data augmentation, we design a universal physical-aware density-resampling data augmentation (PDDA) method to mitigate the performance loss stemming from diverse point densities. 
From the learning methodology viewpoint, we develop a multi-task learning for 3D object detection: during source training, besides the main standard detection task, we leverage an auxiliary self-supervised 3D scene restoration task to enhance the comprehension of the encoder on background and foreground details for better recognition and detection of objects.
Furthermore, based on the auxiliary self-supervised task, we propose the first test-time adaptation method for domain generalization of 3D object detection, which efficiently adjusts the encoder's parameters to adapt to unseen target domains during testing time, to further bridge domain gaps. 
Extensive cross-dataset experiments covering ``Car'', ``Pedestrian'', and ``Cyclist'' detections, demonstrate our method outperforms state-of-the-art SDG methods and even overpass unsupervised domain adaptation methods under some circumstances. 
% The code will be made publicly available.
The code is released at \href{https://github.com/xingyu-group/3D-Density-Resampling-SDG}{https://github.com/xingyu-group/3D-Density-Resampling-SDG}.
\end{abstract}

\section{Introduction}
\label{sec:intro}

\begin{figure}[t]
  \centering
   \includegraphics[width=0.99\linewidth]{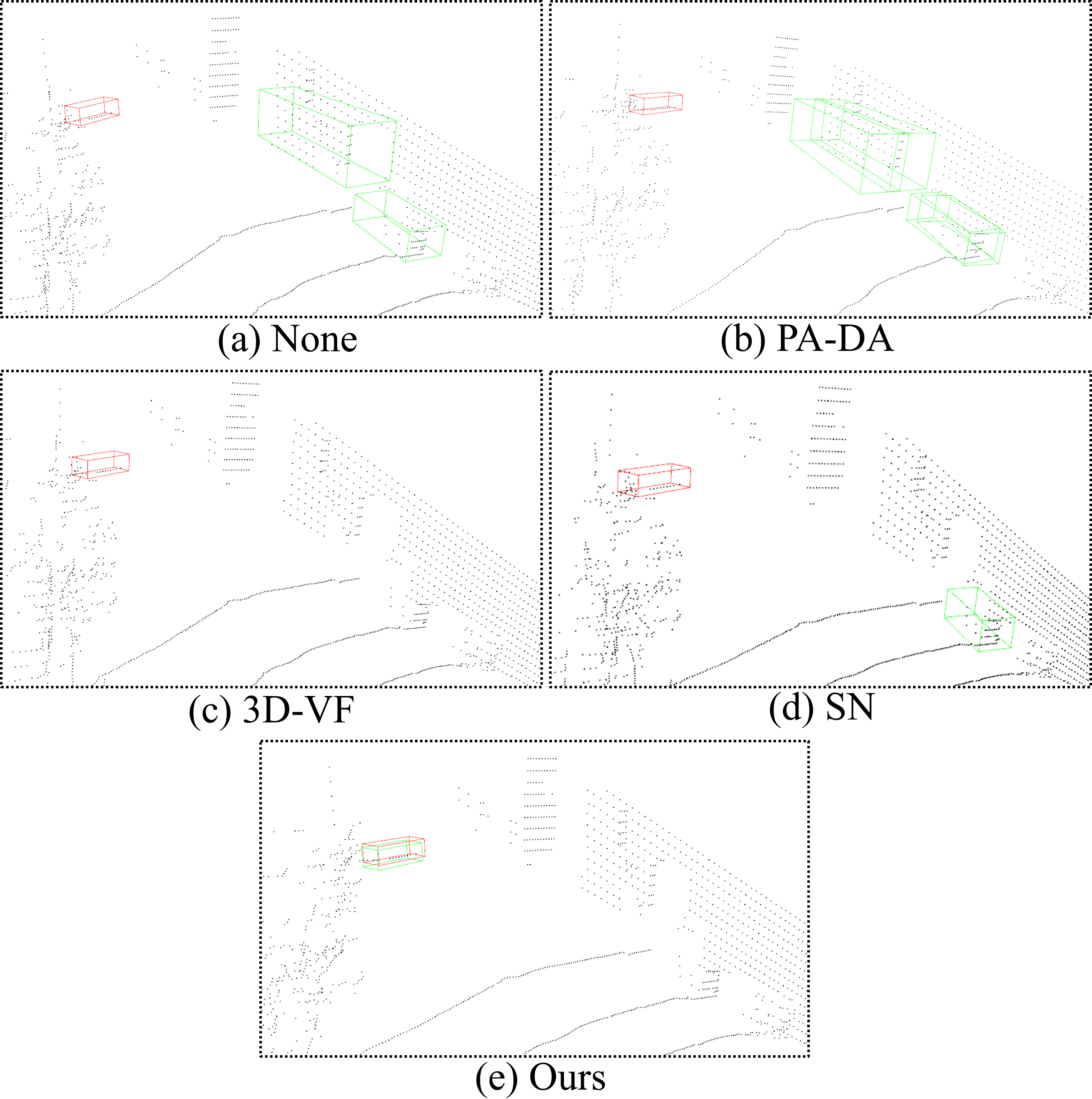}
   \caption{Detection results \wrt Waymo $\rightarrow$ NuScenes, where the red boxes are ground-truth 3D boxes and green ones are detected 3D boxes. Our method achieves better performance than other SDG methods, \textit{PA-DA}~\cite{choi2021part} and \textit{3D-VF}~\cite{lehner20223d}, and even UDA method \textit{SN}~\cite{wang2020train} (refers to Table~\ref{tab: main_comparision_DG_UDA} for statistical details).}
   \label{fig: waymo_to_nuscenes_detection}
\end{figure}

% Lidar-based point cloud的3d object detection是自动驾驶领域一个重要的任务。但在面对不同于training时的unseen domain 数据时，3d object detection会发生严重的性能损失。针对这个问题，domain generalization … (定义)。类似的，uda，不同点。

LiDAR-based 3D object detection is a crucial task in real-world applications, such as autonomous driving~\cite{arnold2019survey, qian20223d} and UAV sensing~\cite{ye2020sarpnet, hammer2018lidar}. Deep learning-based models ~\cite{arnold2019survey, qian20223d, ye2020sarpnet, yan2018second, shi2019pointrcnn} are the mainstream for 3D object detection. However, when these models encounter data with domain gaps (\eg, adverse weather~\cite{hahner2022lidar, kilic2021lidar, hahner2021fog}, diverse laser scanning densities~\cite{hu2023density}, and shifted object sizes~\cite{wang2020train}), their performance declines significantly. 
To tackle it, unsupervised domain adaptation (UDA)~\cite{wang2020train, yang2021st3d++, hu2023density} focuses on adapting the model to a new domain distribution during training with the help of unlabeled target data. 
Targeting a more general case, domain generalization (DG)~\cite{volpi2019addressing, muandet2013domain, iwasawa2021test} endeavors to enhance the model's generalizability with no access to target data during model training.

% 目前3d dg的研究（两类）。
% mdg可以提供diverse的info, to bridge the domain gaps by unseen domain
% 考虑到real-world domain信息的high varierty， multi-domain的数据acquiring和模型training是高成本。对此我们focus到 更通用的single-domain generalization（SDG），即sdg定义。sdg在2D广泛的研究，但在3d pointclouds，尤其是object detection，研究较少。

The approaches of domain generalization in 3D object detection can be primarily categorized into multi-domain generalization and single-domain generalization. Multi-domain generalization (MDG)~\cite{wu2023towards, soum2023mdt3d} utilizes diverse source domain knowledge to bridge the unseen target domain gaps. Given the high diversity of real-world domains, 3D data collection and label annotation for multiple source domains is costly. In this paper, we focus on the more universal yet more challenging single-domain generalization (SDG) setting, where the model learns from a single source domain only. % to enhance its generalization to unseen target domains. Single-domain generalization's 
The primary challenge of SDG lies in how to rely on limited single-domain information to enable a model to achieve domain-invariant learning and therefore perform robustly on diverse unseen domains. %Currently, single-domain generalization~\cite{zhou2022domain} has gathered significant attention in various 2D image tasks (\eg, classification~\cite{wang2021learning}, detection~\cite{vidit2023clip}, and segmentation~\cite{ouyang2022causality}). However, in the realm of the 3D point cloud, especially for object detection tasks, the exploration of single-domain generalization remains relatively rare.
Though single-domain generalization~\cite{zhou2022domain} has gathered significant attention in various 2D image tasks (\eg, classification~\cite{wang2021learning}, detection~\cite{vidit2023clip}, and segmentation~\cite{ouyang2022causality}), it's still under-explored in the context of the 3D point cloud, especially for object detection. Prior works in SDG of 3D object detection like \textit{PA-DA}~\cite{choi2021part} and \textit{3D-VF}~\cite{lehner20223d} usually utilize data augmentation to synthesize more 3D data, aiming to eliminate domain-dependent information in model learning. Yet, no effort has been reported from the learning methodology perspective.

\begin{figure}[t]
  \centering
   \includegraphics[width=1\linewidth]{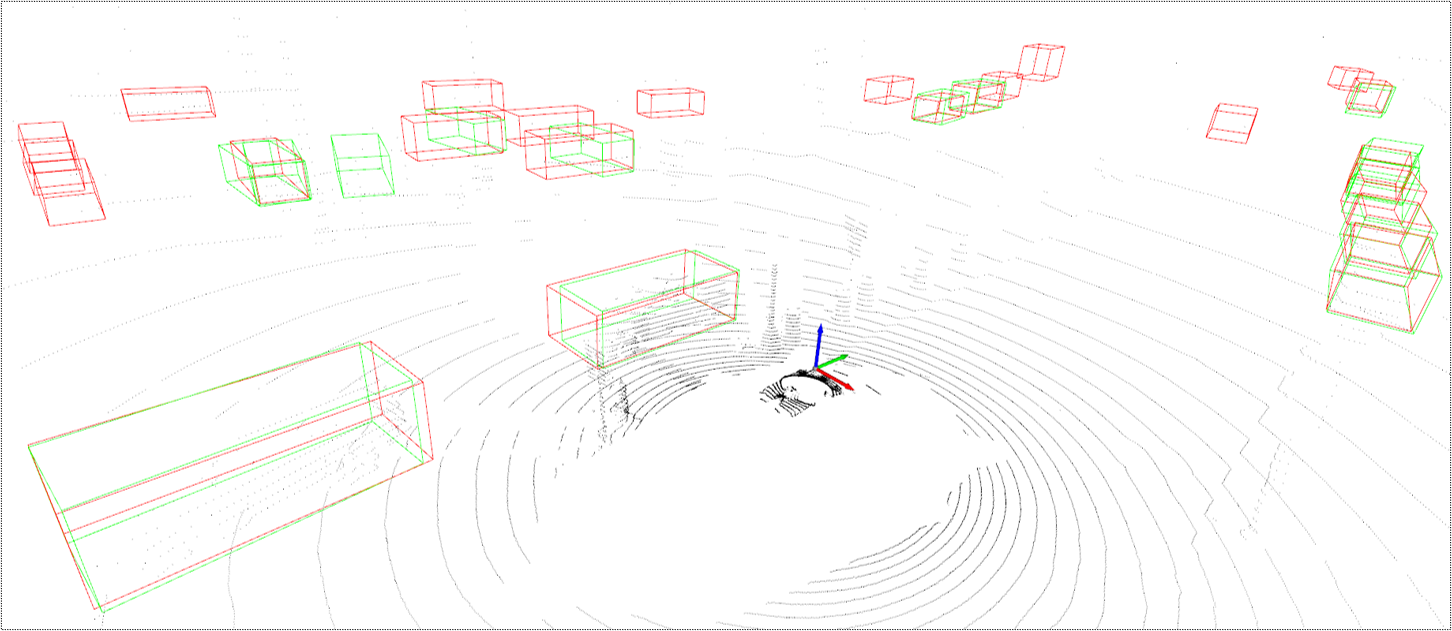}
   \caption{Intra-domain detection by VoxelRCNN~\cite{deng2021voxel} with voxel-based backbone on NuScenes. Due to the blockage by front objects and various distances, cars with sparse scanning are hard to detect. (red boxes are ground-truth 3D boxes and green ones are detected 3D boxes)}
   \label{fig: intra_nuscenes_detection}
\end{figure}

In this work, we tackle the SDG problem from both data augmentation perspective and multi-task learning strategy. The former bridges the domain gap mainly introduced by various point densities, and the latter facilitates representation learning through 3D scene restoration and test-time adaptation. 
Specifically, our data physical-aware data augmentation stems from our observation in intra-domain 3D object detection: \textit{the low point density is highly correlated to object miss-detection}. As demonstrated in Figure~\ref{fig: intra_nuscenes_detection}, object occlusion~\cite{xu2022behind} causes significant local point density reduction and various distances between objects and the laser sensor~\cite{hu2022point} also cause variations in the point density on imaged objects. In inter-domain cases, further considering adverse weather (\eg, rain~\cite{kilic2021lidar}, snow~\cite{hahner2022lidar}, and fog~\cite{hahner2021fog}) that may cause the reduced intensity of reflected light resulting in sparse laser scans and the diverse 3D sensors with different scanning beam layers, we hypothesize that the issue introduced by point density variations becomes more pronounced in inter-domain object detection. In this regard, we design a universal physical-aware density-resampling data augmentation (PDDA) to mitigate the performance loss stemming from diverse point densities. Unlike prior augmentation methods for domain generalization of 3D object detection like \textit{PA-DA}~\cite{choi2021part} and \textit{3D-VF}~\cite{lehner20223d} that modify point density augmentation in a localized and random manner, our PDDA re-samples the point clouds following real-world imaging physical constraints, thus better accounting for different density patterns. %with following the real-world uniform distribution of diverse point layers.

From the learning methodology viewpoint, we propose a multi-task learning for point-cloud-based 3D object detection. Specifically, during source training, besides the standard detection task, we design an auxiliary self-supervised task to restore the globally uniformly masked points by density downsampling. Recently, some research~\cite{boyce1989effect, zheng2022hyperdet3d} has shown that background information is also crucial in object recognition and detection. By working with the standard detection task, our self-supervised task helps the encoder to better comprehend the background and foreground details in point clouds, which benefits the recognition and detection of objects. 
Furthermore, during the testing time, leveraging the auxiliary self-supervised task for 3D scene restoration, we design the restoration loss to measure the quality of restoration on the target domain data. Based on the loss, we efficiently adjust the encoder's parameters with lightweight optimization to adapt the encoder to unseen target domain shifts, thereby further bridging domain gaps. In summary, this work makes the following contributions:
\begin{itemize}
    \item To tackle the erroneous object detections caused by diverse point densities, we design a straightforward yet effective universal physical-aware density-resampling data augmentation method for the source training, which increases the model's resistance to potential domain shifts caused by diverse point densities.
    \item To enhance the model's scene comprehension, we design a multi-task learning with 3D scene restoration to boost the recognition and detection of objects during source training. Incorporating the 3D scene restoration task, we further utilize the lightweight parameter updates to adapt the model to unseen target domains during testing time, to further bridge domain gaps.
    \item Extensive cross-dataset experiments covering ``Car'', ``Pedestrian'', and ``Cyclist'' detections demonstrate our model outperforms the state-of-the-art DG methods and even overpass UDA methods under some circumstances.
\end{itemize}

\section{Relate Work}
\label{sec:formatting}

\subsection{Point-Cloud-Based 3D Object Detection}

% 介绍detection。 detection的input representation分为两类。 voxel-based介绍，优缺点。x方法s。 point-based ...。 此外还有mix使用两种的point-voxel-based， 优缺点。我们主要focus到...。

Point-cloud-based 3D object detection~\cite{arnold2019survey, qian20223d, yan2018second, deng2021voxel, shi2019pointrcnn, shi2020pv} aims to locate and classify objects of interest in point clouds. Existing detection methods can be categorized into voxel-based and point-based. Voxel-based methods \cite{yan2018second, lang2019pointpillars, deng2021voxel, mao2021voxel} voxelize point clouds into grid-like 3D images and use sparse convolution to extract abstract features for object detection, which are computationally efficient but may lose fine-grained details due to voxelization. The point-based methods \cite{shi2019pointrcnn, shi2020point} extract abstract features directly from raw points, preserving fine-grained details but possessing relatively heavy computation. As a trade-off, some point-voxel-mixed methods \cite{shi2020pv, shi2023pv} combine raw points and voxels to balance representation learning and computation efficiency. In this article, we mainly focus on voxel-based object detection.

\subsection{Domain Generalization on 2D/3D Object Detection}
Domain generalization~\cite{volpi2019addressing, muandet2013domain, iwasawa2021test} aims to bridge domain shifts unseen during the source training but may encountered in the target testing. In 2D image tasks, domain generalization methods have been well explored and include data augmentation~\cite{volpi2019addressing, volpi2018generalizing}, domain alignment learning~\cite{muandet2013domain, motiian2017unified}, meta-learning~\cite{li2018learning, balaji2018metareg}, test-time adaptation~\cite{iwasawa2021test, chen2023improved}, \etc. In the realm of 3D object detection, existing works ~\cite{wu2023towards, soum2023mdt3d} mainly focus on multi-domain generalization. Wu \etal in \cite{wu2023towards} propose to align extracted features of detectors individually trained on multiple source domains to learn domain-invariant features. Soum-Fontez  \etal in \cite{soum2023mdt3d} utilize label re-annotation and cross-dataset instance injection to mitigate the performance degradation.
However, regarding single-domain cases, although well-explored in 2D image detection~\cite{wu2022single, vidit2023clip, zhou2022domain}, there has been relatively rare research in 3D object detection. Lehner \etal in \cite{lehner20223d} propose the adversarial augmentation to deal with rare-shape or broken cars. Choi \etal in \cite{choi2021part} combine 5 basic augmentations to build the part-aware data augmentation to handle noise and locally missing points. Unlike prior works solely focusing on data augmentation, our work not only introduces a novel physical-aware data augmentation method, but also contributes a new multi-task learning framework. %strategy in methodology.

\subsection{Test-Time Adaptation in Domain Generalization}
Testing-time adaptation ~\cite{chen2022contrastive, liang2023comprehensive} refers to conducting adjustments or refinements to a model's parameters or inferences during the testing or inference phase. Current methods include pseudo-labeling~\cite{liang2020we, chen2022self}, consistency training~\cite{wang2020tent}, self-supervised learning~\cite{chen2022contrastive, mirza2023mate}, \etc. 
Recently, the increasing attention has landed on using test-time adaptation to tackle the domain generalization problem in 2D tasks~\cite{chen2023improved, iwasawa2021test, liu2022single, chen2022contrastive, liang2020we, chen2022self} %, namely generalizing the model to unlabeled target domain data to bridge the domain gaps during testing.Existing methods for test-time training are primarily used in 2D tasks~\cite{chen2022contrastive, liang2020we, chen2022self} 
and 3D classification~\cite{wang2020tent, mirza2023mate}.  To the best of our knowledge, our work is the first to apply the test-time adaptation to the SDG for 3D object detection.
\section{Problem Formulation}
3D object detection aims to detect objects of interest in 3D point clouds. A frame of point cloud $\textbf{X}$ is a set of points $p = [x^p, y^p, z^p]$, where $[x^p, y^p, z^p]$ denotes the 3D Cartesian coordinates of points. The object detection can be formulated as $f(\textbf{X}) = \{\textbf{b}_i\}_{i=1}^{N_{obj}}$, where $f(\cdot)$ denotes the detection model and $N_{obj}$ is the number of detected 3D boxes in $\textbf{X}$. $\textbf{b}_i$ denotes the $i^{th}$ detected box, which contains the location $[x_i, y_i, z_i]$, the dimension $[w_i, h_i, l_i]$, the heading angle $\theta_i$, the classification label $c_i$, and the confidence score $s_i$.

Domain generalization for 3D object detection aims to generalize the model $f(\cdot)$ trained on a well-labeled source-domain dataset $\{\textbf{X}_j^s, \textbf{B}_j^s\}_{j=1}^{N^s}$ to a unseen target domain $\{\textbf{X}_j^t,\textbf{B}_j^t\}_{j=1}^{N^t}$, where the superscript $s$ and $t$ denote the source and target domain respectively. Correspondingly, $N^s$ and $N^t$ denote the number of point clouds, and $\textbf{B}_j^s$ and $\textbf{B}_j^t$ denote the sets of 3D boxes. Note that in domain generalization, only the source-domain data is available for model training. The target-domain data is accessible during model evaluation/deployment only.

%The source domain is the labeled dataset $\{\textbf{X}_j^s, \textbf{B}_j^s\}_{j=1}^{N^s}$ where $N^s$ denotes the number of point clouds and $\textbf{B}_j^s$ denotes a set of 3D boxes in the $j_{th}$ point cloud, while the target domain is the unlabeled dataset $\{\textbf{X}_j^t\}_{j=1}^{N^t}$, where $s$ and $t$ denotes the source and target domain respectively. To bridge the domain gaps in the target domain, we propose our single-domain generalization method as in Figure \ref{fig: framework}, which mainly consists of three components: 

\section{Methodology}
\begin{figure*}[t]
  \centering
   \includegraphics[width=0.88\textwidth]{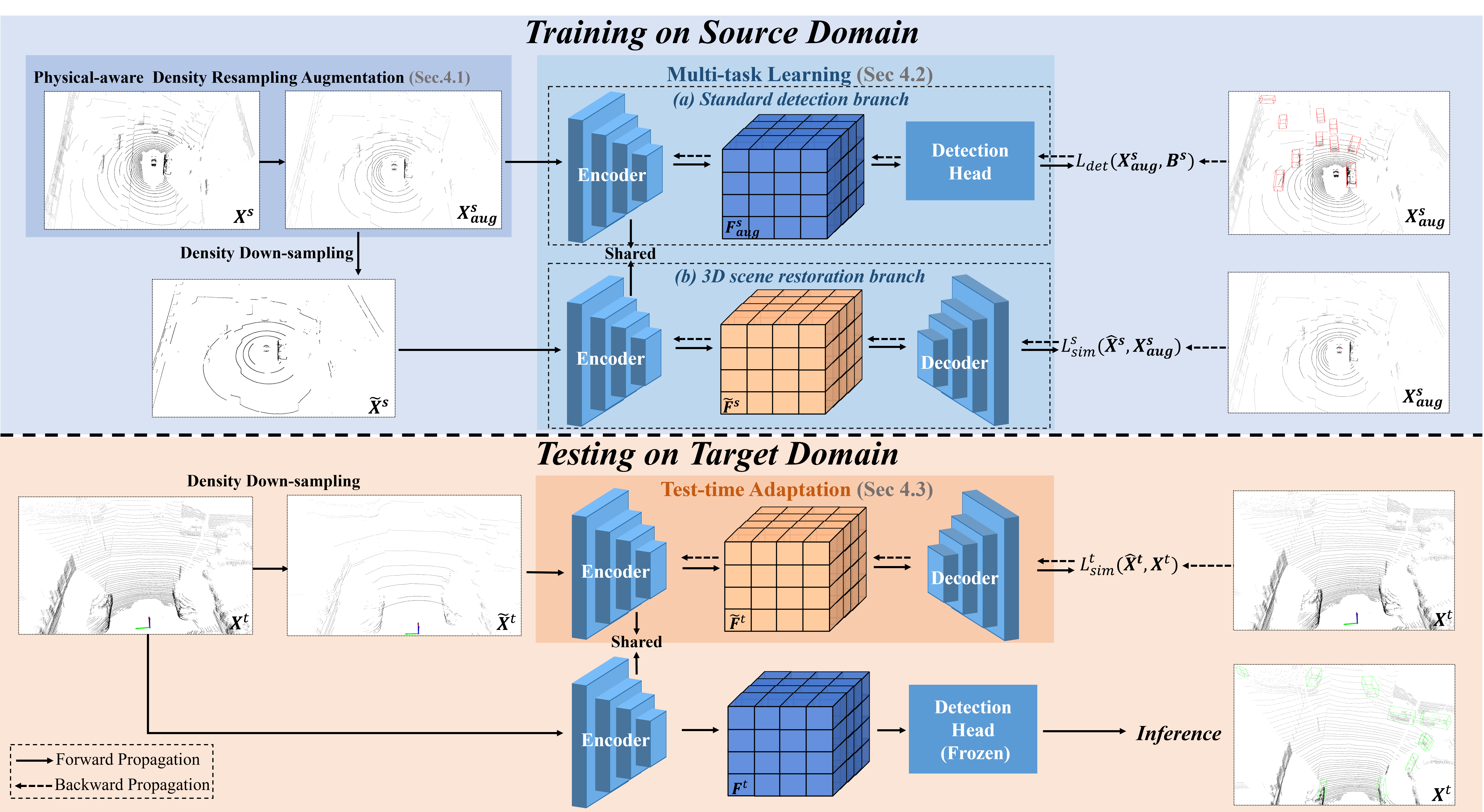}
   \caption{Pipeline of our proposed DG method. During training on the source domain, the training sample is augmented with density re-sampling, which is then used to train the multi-task model for (a) standard detection and (b) 3D scene restoration from its down-sampled version. During Testing on the target domain, given a query data, self-supervised scene restoration is conducted on the corresponding density-downsampled version for lightweight model update. Then the updated encoder works together with the frozen detection head for the final prediction. In this figure, source and target samples are from NuScenes~\cite{caesar2020nuscenes} and KITTI~\cite{geiger2013vision}, respectively.}%Pipeline of our proposed DG method. We take the NuScenes $\rightarrow$ KITTI task and the density down-sampling of PDDA for example. During training on the source domain, the given NuScenes sample is first augmented with density down-sampling. In multi-task learning, the augmented sample is used for (a) standard detection, and also is down-sampled and used for (b) 3D scene restoration. During Testing on the target domain, the given KITTI sample is first density-downsampled, and after limited iterations, the updated encoder together with the frozen detection head outputs the final prediction.}
   \label{fig: framework}
\end{figure*}

Figure \ref{fig: framework} presents the system diagram of the proposed DG solution on 3D point cloud data. During training on the source domain, the point-cloud data is augmented by our PDDA method. Then the augmented sample is fed into the multi-task learning scheme, where the auxiliary self-supervised task (i.e. 3D scene restoration) promotes the encoder to extract features for scene comprehension. During testing on the target domain, we utilize the auxiliary 3D scene restoration loss to update parameters in the encoder-decoder path to adapt the encoder to unseen target domains, to further bridge the domain gap. The test-time adapted encoder together with the frozen detection head is then exploited for the final object detection.

%\begin{itemize}
%    \item To tackle the erroneous object detections caused by diverse object densities in intra-domains and inter-domains, we design a straightforward yet effective universal physical-aware density-resampling data augmentation method during the source training, detailed in Section~\ref{sec: uni_den_aug}.
%    \item To help the model better sense background and foreground information, we develop a density-downsampling-based self-supervised training task during source training, detailed in Section~\ref{sec: den_down_self_train}.
%    \item By means of the designed self-supervised training, we design the test-time training to enhance the model's resistance to unseen target domain gaps during testing, detailed in Section~\ref{sec: test_time_train}.
%\end{itemize}

\subsection{Physical-Aware Density-Resampling Data Augmentation}
\label{sec: uni_den_aug}

% 在点云成像里，point density是影响物体认知的关键因素。常见的，domain之内的不同距离和物体遮挡和domain之间的天气都是可以直接影响成像点云的点density。此外，不同数据集所使用的不同lidar sensor，也会造成vertical 点密度的不同，如xxx所示。针对此，xx提出了random beam re-sampling 方法，即通过随机mask掉或者interpolate beam layer去增加pre-training阶段的density多样性。但随机sampling很难simulate出实际的不同lidar sensor中vertical beam layer的均匀分布。对此，我们设计了physical-aware的beam resampling方法去simulate 实际的不同密度的点云分布。

%Point density is an essential factor that affects the performance of the 3D detection model. Commonly, various distances of objects \wrt LiDAR sensors and inter-object occlusion, as well as different weather conditions, directly influence the point density in the captured point cloud. Besides, different types of LiDAR sensors may also result in variations in point density, which further degrades generalization ability on a unseen domain. Note that point density variations introduced by distance and sensors follows different physical constraints. The former one leads to sparse sampling in horizontal with propotional to the distance, while the latter may various density in verticle direction.  %, as shown in Table\ref{tab: dataset_density}. 

Point density plays a crucial role in 3D object detection. It is affected by several factors, including the distances between objects in relation to LiDAR sensors, inter-object occlusion, and different weather conditions. Furthermore, various types of LiDAR sensors can introduce variations in point density, which can negatively impact the model's ability to generalize to unseen environments. 
% It's important to note that the variations in point density due to distance and sensor types follow distinct physical constraints. The former results in sparser sampling horizontally, which is proportional to the distance, while the latter can lead to varying uniform-distributed densities in the vertical direction.
It's important to note that the variations in point density due to various factors %sensor types
may follow distinct physical constraints. For example, point density is inversely proportional to the imaging distance.
Furthermore, for spinning sensors, different beam layers of the sensor can lead to varying uniform-distributed densities in the vertical direction. 
To address the domain-specific bias arising from variations in point densities, Hu \etal \cite{hu2023density} proposed a random beam re-sampling (RBRS) method. %, which involves randomly masking or randomly interpolating beam layers to increase density diversity during the pre-training phase. However, random sampling makes it hard to simulate the real-world vertically uniform distribution of beam layers regarding LiDAR sensor scanning. 
Nonetheless, this solely randomly layer-sampling strategy does not adhere to the previously mentioned physical constraints, resulting in suboptimal outcomes. In this study, we design a physical-aware density-resampling data augmentation (PDDA) method to better simulate the real-world point cloud distribution, accounting for different density patterns.

%我们假设获取source dataset的lidar sensor相关的beam layer数量。对于一个point cloud, 我们将其cartesian coordinates转化spherical coordinates。

Given the point clouds of the source dataset with $M$ beam layers of LiDAR scanning, we first convert the Cartesian coordinates $[x, y, z]$ into spherical coordinates $[\gamma, \Theta, \Phi]$:
\begin{equation}
    \gamma=\sqrt{x^2 + y^2 + z^2}, \Theta=\arctan{\frac{y}{x}}, \Phi=\arccos{\frac{z}{\gamma}}. 
\end{equation}
Considering the spinning beams commonly used in autonomous driving, we gather vertical angles $\{\Phi\}$ of points. After removing outliers of $\{\Phi\}$ (out of $3.1\times$ standard deviation), we obtain the range of vertical angles $[\min\{\Phi\}, \max\{\Phi\}]$. Then we uniformly divide the range into $M$ bins $[\Phi_k, \Phi_{k+1}]$ ($k\in\{1, 2, \cdots, M\}$) and label points according to which bin the points land on. Then we perform our density-resampling data augmentation on source data by comprehensively considering physical-aware down-sampling and up-sampling. 

To uniformly \textit{down-sample} the density of the point cloud into a lower density, we keep one out of every $C$ bins of points. A higher value of $C$ means a more severe reduction in density.  
Additionally, we remove points with a probability of $P$ to simulate the global loss of laser reflections. 
To \textit{up-sample} the density of the point cloud into a higher density, we adopt efficient linear interpolation to obtain new points. Specifically, new points will be interpolated by original points within two neighboring bins: 
\begin{equation}
\begin{aligned}
    &\eta^{new}_s=\lambda\eta_k + (1-\lambda)\eta_{k+1}\\ 
    &\text{s.t. } s\in\{1, 2,\cdots, S-1\},  \lambda=\frac{s}{S},
\end{aligned}
\end{equation}
where $\eta \in \{\gamma, \Theta, \Phi\}$ represents the spherical coordinate of points; $k$ refers to points with $\Phi$ belonging to the $k_{th}$ bins; $s$ indicates the new points of $s_{th}$ interpolated beam layer; $\lambda$ is the interpolation ratio; the integer $S$ ($S>2$) is the interpolation factor and $S-1$ is the number of newly interpolated beam layers. 

Considering low-density is relatively more detrimental to object detection than high-density, we select two down-sampling operations with $C\in\{2, 3\}$ (\ie, \textit{$2\times$down-sampling} and \textit{$3\times$down-sampling}) and one up-sampling operation with $S=2$ (\textit{$2\times$up-sampling}). 
At last, we design the universal density-resampling augmentation method operating on source training data, which randomly conducts one of \{\textit{$2\times$down-sampling, $3\times$down-sampling, no-sampling, $2\times$up-sampling }\} on each source point cloud to simulate potential various densities on target domains. 
More visualizations are shown in Section~\ref{sec: suppl_PDDA} in the supplementary.

\subsection{Multi-Task Learning with Density-Resampling}
\label{sec: den_down_self_train}

As shown in Figure~\ref{fig: framework}, our multi-task learning scheme consists of the main standard detection task and the auxiliary self-supervised task. As we design, the auxiliary self-supervised task restores the globally uniformly masked points by density downsampling. Recently, some research~\cite{boyce1989effect, zheng2022hyperdet3d} has shown that background information is also crucial in object recognition and detection. 
% Besides interacting with foreground points, background points also imply the noise and density properties of point clouds. 
Through our self-supervised task, the restoration of the 3D scene helps the encoder to better comprehend the background and foreground details of the scene, which benefits the model's object recognition and thereby object detection. 

\noindent \textbf{Main standard detection task.} By the PDDA in Section \ref{sec: uni_den_aug}, we obtain the augmented point cloud $\textbf{X}^s_{aug}$ from the original point cloud $\textbf{X}^s$. Then, we feed $\textbf{X}^s_{aug}$ into the encoder to obtain abstract features $F_{aug}$ by $F_{aug} = f_{\theta_{E}}(\textbf{X}^s_{aug})$, where $\theta_{E}$ is the parameter of the encoder. Following the setting in \cite{deng2021voxel}, we calculate the standard detection loss $\mathcal{L}_{det}$:
\begin{eqnarray}
    \mathcal{L}_{det}(\textbf{X}^s_{aug}, \textbf{B}^s; \theta_E, \theta_H) = f_{\theta_H}(F_{aug}),
\end{eqnarray}
\noindent where $\theta_{H}$ is the parameter of the detection head.

\noindent \textbf{Auxiliary self-supervised task.} By density down-sampling in Section \ref{sec: uni_den_aug}, we further down-sample $\textbf{X}^s_{aug}$ into $\tilde{\textbf{X}}^s$. Then, we feed $\tilde{\textbf{X}}^s$ into the encoder to obtain the abstract feature $\tilde{F}$ by $\tilde{F} = f_{\theta_{E}}(\tilde{\textbf{X}^s})$. 
For 3D scene restoration, we use the decoder (symmetrical to the encoder) to restore the point cloud $\hat{\textbf{X}}^s$ from $\tilde{F}$, namely $\hat{\textbf{X}}^s=f_{\theta_D}(\tilde{F})$ where $\theta_D$ is the parameter of the decoder. To align $\hat{\textbf{X}}^s$ with $\textbf{X}^s_{aug}$, we first adopt the mean squared error (MSE) to calculate the  $\mathcal{L}_{mse}^s$.
\begin{eqnarray}
\centering
    &\mathcal{L}_{mse}^s(\tilde{\textbf{X}}^s, \textbf{X}^s_{aug}; \theta_E, \theta_D) = {\lVert \hat{\textbf{X}}^s-\textbf{X}^s_{aug} \rVert}_2^2.
\end{eqnarray}
To improve semantic similarity, we also leverage pre-trained PointNet++~\cite{qi2017pointnet++} to acquire multi-scale abstract features of $\hat{\textbf{X}}^s$ and $\textbf{X}^s_{aug}$ and calculate the perceptual loss~\cite{yang2018low}: 
\begin{eqnarray}
\centering
\begin{split}
    \mathcal{L}_{pcp}^s (\tilde{\textbf{X}}^s, \textbf{X}^s_{aug}; \theta_E, \theta_D) & =  \\ \sum_{i=1}^3 &\mathcal{L}_{cos}(\text{PNet}_i(\hat{\textbf{X}}^s), \text{PNet}_i(\textbf{X}^s_{aug})),
\end{split}
\end{eqnarray}
\noindent where the cosine similarity loss $\mathcal{L}_{cos}(\textbf{A}, \textbf{B}) = 1- \frac{\textbf{A}\cdot\textbf{B}}{\lVert \textbf{A}\rVert \lVert \textbf{B}\rVert}$ and $\text{PNet}_i(*) (i\in\{1, 2, 3\})$ outputs the flattened feature of the $i_{th}$ block of the PointNet++ encoder. Thus, the compound loss of the self-supervised training $\mathcal{L}_{self}^s$,
\begin{equation}
    \mathcal{L}_{self}^s = \mathcal{L}_{mse}^s + \lambda_1\mathcal{L}_{pcp}^s,
\end{equation}
is used to restore the down-sampled point cloud for a better comprehension of the 3D scene.

% Regarding the source training procedure, we provide two training modes: the \textbf{end-to-end training} jointly the model by means of $\mathcal{L}_{det}$ + $\lambda_2\mathcal{L}_{self}^s$, while the \textbf{two-stage training} first conduct standard detection training to update the encoder and the detection head by $\mathcal{L}_{det}$ and then freeze the encoder to update the parameters of the decoder and statistical parameters (\ie, the mean and the standard deviation) of BatchNorm layers of the encoder by $\mathcal{L}_{self}^s$.

Regarding the multi-task learning scheme, we adopt the two-stage training as in Figure~\ref{fig: framework}: first, we conduct standard detection training to update the encoder and the detection head by $\mathcal{L}_{det}$ and then utilize the self-supervised training to update the parameters of the decoder by $\mathcal{L}_{self}^s$. During the self-supervised training, we freeze the major trainable parameters of the encoder to ensure the output features aligned with the detection head and only update statistical parameters (\ie, the mean and the standard deviation) of BatchNorm layers to adapt the encoder to diverse feature styles given various densities. 

\subsection{Test-Time Adaptation with Self-Supervised 3D Scene Restoration}
\label{sec: test_time_train}
Prior research in the field of 2D domain generalization has shown that test-time adaptation can be a highly effective approach to mitigate the disparities between source and target domains~\cite{chen2023improved, iwasawa2021test, liu2022single}. It utilizes lightweight test-time optimization to adjust the model's parameters during the testing. Nevertheless, this strategy has yet to be explored in the context of domain generalization of 3D point-cloud-based object detection.

%Test-time training is raising increasing attention on tackling domain generalization issues in 2D tasks~\cite{chen2023improved, iwasawa2021test, liu2022single}. Specifically, test-time training utilizes rapid test-time training to adjust the model's parameters during the testing. In this way, the model could be effectively adapted to the novel target domain gaps that are unseen in the source training phase. Recently, the test-time training on domain generalization has raised increasing attention in 2D image tasks~\cite{chen2023improved, iwasawa2021test, liu2022single}. In this paper, we propose the first test-time training method working on the domain generalization of 3D point-cloud-based object detection, which aims to adapt the detection model to the unseen target domain. 

Given the query data from an unseen domain, the effectiveness of test-time adaptation hinges on the proper construction of self-supervised learning. In this study, we make use of the auxiliary 3D scene restoration task proposed in our multi-task learning to adapt the parameters of the encoder to the target domain.
Specifically, given a point cloud $\textbf{X}^t$ in the target domain, we first down-sample $\textbf{X}^t$ with the proposed PDDA augmentation, obtaining $\tilde{\textbf{X}^t}$. Then $\tilde{\textbf{X}^t}$ is interpolated by the encoder and decoder trained on the source data and the corresponding restoration loss
\begin{eqnarray}
\centering
    &\mathcal{L}_{mse}^t(\tilde{\textbf{X}}^t, \textbf{X}^t; \theta_E, \theta_D) = {\lVert f_{\theta_D}(f_{\theta_E}(\tilde{\textbf{X}}^t))-\textbf{X}^t \rVert}_2^2
\end{eqnarray}
is used to update $\theta_E, \theta_D$, improving the encoder's comprehension of the query scene. One may notice that the perceptual loss in (6) is omitted in our test-time adaptation, even though it leads to a marginal improvement in test-time optimization. This decision was made due to the significant computational latency introduced by the perceptual loss. We show this trade-off in Table~\ref{tab: Suppl_effect_pcp_loss} in the Supplementary. 

During testing time, for each query data, we reset the parameters $\theta_E$ and $\theta_D$ to the initial state produced in the source training phase. We then iteratively update them a specific number of times, denoted as $N_{iter}$, to minimize the point-cloud restoration loss $\mathcal{L}_{mse}^t$. This approach ensures both high computational efficiency and detection performance improvement. After undergoing $N_{iter}$ updates using the data from point cloud $\textbf{X}^t$, the encoder is combined with the detection head to produce the final detection result.

\section{Experiment}

\subsection{Experiment Settings}

\begin{table}[t]
\centering
\caption{The overview of 3D point cloud datasets}
\label{tab: dataset_density}
\resizebox{\linewidth}{!}{%
\begin{tabular}{@{}cccc@{}}
\toprule
\textbf{Dataset} & \textbf{Frame} & \textbf{LiDAR Sensor} & \textbf{Vertical View}  \\ \midrule
Waymo\cite{sun2020scalability} & 200K & \begin{tabular}[c]{@{}c@{}}1 x 64 spinning-beam + \\ 4 x 200 fixed-beam\end{tabular} & {[}-17.6, 2.4{]}  \\
KITTI\cite{geiger2013vision} & 15K & 1 x 64 spinning-beam & {[}-24.9, 2.0{]}  \\
NuScenes\cite{caesar2020nuscenes} & 40K(Labeled) & 1 x 32 spinning-beam & {[}-30.0, 10.0{]}  \\ \bottomrule
\end{tabular}%
}
\end{table}

\noindent \textbf{Datasets and metrics.} We select select three widely-recognized datasets of autonomous driving: KITTI~\cite{geiger2013vision}, Waymo~\cite{sun2020scalability}, NuScenes~\cite{caesar2020nuscenes}, as shown in Table~\ref{tab: dataset_density}. 
% KITTI is a 3D object detection dataset collected in Germany and contains 3,712 training and 3,769 frames of point clouds collected by a spinning 64-beam LiDAR. Waymo is a large-scale 3D object detection dataset in the USA and comprises $\sim$158K training frames and $\sim$40K validation frames collected by one 64-beam spinning LiDAR and four 200-beam fixed LiDARs. For Waymo point clouds collected continuously over time, we evenly sample $20\%$ of the data for experiments. The dataset NuScenes collected in Boston and Singapore contains $\sim$28K training labeled key-frames and 6,019 validation labeled key-frames. 
KITTI contains $\sim$15K frames of point clouds collected by a spinning 64-beam LiDAR. Waymo comprises $\sim$200K frames collected by one 64-beam spinning LiDAR and four 200-beam fixed LiDARs. For Waymo samples collected continuously over time, we evenly sample $20\%$ of them for experiments. NuScenes contains $\sim$40K labeled key-frames. 
Three datasets present shifted distributions caused by different LiDAR types, geographic collection locations, inconsistent object sizes, \etc. Following conventional settings in ~\cite{hu2023density, yang2021st3d, yang2021st3d++}, we evaluate models under cross-dataset settings of NuScenes $\rightarrow$ KITTI, Waymo $\rightarrow$ NuScenes, and Waymo $\rightarrow$ KITTI.
We select ``Car'' (also ``Vehicle'' in Waymo), ``Pedestrian'' and ``Cyclist'' (also ``bicycle'' in NuScenes) for detection. For the fair evaluation across datasets, we take the average precision (AP) of 40 recalls and the mean AP (mAP) averaging on all object classes on both 3D and BEV views as the evaluation metrics. The IoU thresholds are 0.7 for ``Car'' and 0.5 for ``Pedestrian'' and ``Cyclist''. Note that, for KITTI, we record the average AP at all difficulty levels (\ie, Easy, Moderate, and Hard). 

\noindent \textbf{Implementation details.} We evaluate all DG and UDA methods based on the powerful detector VoxelRCNN~\cite{deng2021voxel} which uses voxel-based features for both representation extraction and box refinement. For a fair comparison, we use the single unified VoxelRCNN simultaneously detecting ``Car'', ``Pedestrian'', and ``Cyclist''.  We conduct the experiments on the popular codebase openpcdet~\cite{openpcdet2020} and openpcdet-based Uni3D~\cite{zhang2023uni3d}.  For the source training, we adopt the down-sampling with $C\in\{3, 4, 6\}$ and the one-cycle Adam optimizer with a learning rate of 0.01 during 30 training epochs. Due to unreachable labeled target data, we select the best checkpoint through the validation with labeled source \textit{val} data. For self-supervised learning, we finetune the model for a limited 5 epochs with $\lambda_1=1.0$. We also adopt common data augmentations for all detections, including random world flipping, random world scaling, random world rotation, and ground-truth object sampling. For the test-time training, we adopt the down-sampling with $C\in\{6, 8\}$, the Adam optimizer with a learning rate of 0.001 and set $N_{iter}=5$ for each sample. 
Regarding no prior information on the target domain, we set the number of beam layers as the default 64. 
Using 2$\times$ GeForce RTX-3090/A100, we set the batch size to 4 in source training and 1 in test-time training following the real-world online frame-flow input. 
For the consistent format of point clouds across datasets, we unify the LiDAR coordinate system with the origin on the ground within the range of $[-75.2m, -75.2m, -2m, 75.2m, 75.2m, 4m]$ and the voxel size of $[0.1m, 0.1m, 0.15m]$.  

\noindent \textbf{Compared methods.} (a) \underline{DG methods:} 3D-VF \cite{lehner20223d} leverages the adversarial vector-field-based augmentation to generalize the detector to the rare-shape or broken objects. PA-DA~\cite{choi2021part} utilizes the part-aware data augmentation combining 5 basic augmentation methods (\ie, \{\textit{dropout, swap, mix, sparse, noise}\}) to enable the detector robust to noise and dropout points. (b) \underline{UDA methods:} SN~\cite{wang2020train} leverages data augmentation to de-bias the impact of different object sizes on model generalization. Incorporating random object scaling (ROS), ST3D++~\cite{yang2021st3d++} designs a self-training pipeline to improve the quality of pseudo-labels of unlabeled target-domain data for cross-domain adaptation. The implementation details can be referred to Section~\ref{sec: suppl_experiment_implimentation} in the supplementary.

%\textbf{Data augmentation methods:} Besides the augmentation methods (\ie, PA-DA, 3D-VF, and SN) in Table~\ref{tab: main_comparision_DG_UDA}, we also include the weather-simulating augmentation methods Rain ~\cite{kilic2021lidar} and Fog~\cite{hahner2021fog} which simulate the adverse rain and fog in point clouds. Regarding UDA augmentation methods which require target-domain information, we also select the RBRS~\cite{hu2023density} and ROS~\cite{yang2021st3d++, yang2021st3d} as compared augmentation methods. The former masks or interpolates beam layers to increase the density diversity of inputs and the latter scales object sizes to mitigate the domain shift of object size statistics. 

\subsection{Comparison with SOTA Methods}
\label{sec: main_experiment}

\begin{table*}[ht]
\centering
\caption{Performance comparison of different methods on DG and UDA tasks. The values on both sides respectively represent the APs on BEV and 3D views (\ie, $AP_{BEV}$/$AP_{3D}$). The bold \textbf{values} represent the best performance in DG tasks and the underlined \underline{values} for the best performance in DG + UDA tasks. ``Source-only'' represents the standard detection model trained by source data without DG, UDA, or additional augmentation methods.}
\label{tab: main_comparision_DG_UDA}
\resizebox{0.72\textwidth}{!}
{%
\begin{tabular}{@{}ccccccc@{}}
\toprule
\multicolumn{2}{c}{\textbf{Tasks}} & \textbf{Methods} & \textbf{Car} & \textbf{Pedestrian} & \textbf{Cyclist} & \textbf{mAP} \\ \midrule
\multirow{6}{*}{NuScenes $\rightarrow$ KITTI} & \multirow{4}{*}{DG} & Source-only & 62.69/19.02 & 22.72/18.37 & 20.61/18.13 & 35.34/18.5 \\  
 &  & PA-DA\cite{choi2021part} & 65.09/32.44 & 18.73/14.94 & 18.66/15.91 & 34.16/21.1 \\
 &  & 3D-VF\cite{lehner20223d} & 65.36/29.21 & 24.85/20.87 & 22.13/\underline{\textbf{19.31}} & 37.45/23.13 \\
 &  & \textbf{Ours} & \textbf{73.58}/\textbf{33.11} & \underline{\textbf{30.01}}/\textbf{23.73} & \underline{\textbf{22.93}}/18.62 & \textbf{42.17}/\textbf{25.15} \\ \cmidrule(l){2-7} 
 & \multirow{2}{*}{UDA} & SN\cite{wang2020train} & 70.5/54.78 & 19.71/15.42 & 13.36/10.83 & 34.52/27.01 \\  
 &  & ST3D++\cite{yang2021st3d++} & \underline{83.57}/\underline{64.17} & 29.27/\underline{25.07} & 16.61/16.12 & \underline{43.15}/\underline{35.12} \\ \midrule \midrule
\multirow{6}{*}{Waymo $\rightarrow$ NuScenes} & \multirow{4}{*}{DG} & Source-only & 31.2/19.13 & 10.52/8.39 & 0.75/0.55 & 14.16/9.36 \\ 
 &  & PA-DA\cite{choi2021part} & 29.43/18.06 & 10.84/8.43 & 0.82/0.43 & 13.7/8.97 \\
 &  & 3D-VF\cite{lehner20223d} & 30.17/18.91 & 10.54/7.23 & 0.76/0.78 & 13.82/8.97 \\
 &  & \textbf{Ours} & \underline{\textbf{36.04}}/\underline{\textbf{22.25}} & \underline{\textbf{14.48}}/\textbf{10.56} & \underline{\textbf{1.15}}/\underline{\textbf{0.95}} & \underline{\textbf{17.22}}/\underline{\textbf{11.26}} \\ \cmidrule(l){2-7} 
 & \multirow{2}{*}{UDA} & SN\cite{wang2020train} & 29.32/18.84 & 12.72/\underline{10.57} & 0.72/0.45 & 14.25/9.96 \\  
 &  & ST3D++\cite{yang2021st3d++} & 27.58/20.25 & 11.88/9.44 & 0.01/0.01 & 13.15/9.9 \\ \midrule \midrule
\multirow{6}{*}{Waymo $\rightarrow$ KITTI} & \multirow{4}{*}{DG} & Source-only & 66.65/19.27 & \textbf{66.55}/\textbf{64.00} & 63.04/57.11 & 65.41/\textbf{46.79} \\
 &  & PA-DA\cite{choi2021part} & 65.82/17.61 & 66.40/63.88 & 61.30/56.23 & 64.51/45.91 \\
 &  & 3D-VF\cite{lehner20223d} & 66.72/19.37 & 66.21/63.12 & 62.74/56.44 & 65.22/46.31 \\
 &  & \textbf{Ours} & \textbf{69.9}/\textbf{20.21} & 63.24/62.59 & \underline{\textbf{63.27}}/\underline{\textbf{57.21}} & \textbf{65.47}/46.67 \\ \cmidrule(l){2-7}  
 & \multirow{2}{*}{UDA} & SN\cite{wang2020train} & 72.43/49.34 & \underline{71.08}/\underline{69.35} & 56.00/53.02 & \underline{66.51}/\underline{57.23} \\  
 &  & ST3D++\cite{yang2021st3d++} & \underline{83.59}/\underline{60.63} & 50.18/48.4 & 52.61/47.38 & 62.13/52.14 \\ \bottomrule
\end{tabular}%
}
\end{table*}

\noindent \textbf{Comparison among DG methods.}
For DG settings, our method outperforms the compared methods, as shown in Table~\ref{tab: main_comparision_DG_UDA}. Specifically, for NuScenes $\rightarrow$ KITTI and Waymo $\rightarrow$ NuScenes where significant low-to-high-density and high-to-low-density domain gaps exist, compared with the second-best performance, the mAPs of our method have increased by ratios of 12.60\%/8.73\% and 21.61\%/20.30\% (with mAP increases of 4.72\%/2.02\% and 3.06\%/1.90\%). It demonstrates our method's generalizability to various density-related domain gaps. 3D-VF and PA-DA aim to overcome the bias caused by the rare-shape objects and noisy and locally missing points. However, due to the ignorance of the point layer variation by different sensors, they lead to limited detection performance. For Waymo $\rightarrow$ KITTI with no significant density-related domain gap, our method still improves the detection accuracy on ``Car'' objects where the bias of object size exists considering ``Car'' in KITTI v.s. ``Vehicle'' (including trunks, vans, etc.) in Waymo, which indicates our method's improvement on the model's generalizability to unseen object sizes.      

% UDA方法借助于reachable的target domain上的unlabeled数据和相关knowledge，来使提高detector在target domain数据上的表现。SN旨在将object size normalize 到target domain上，而ST3D++借助ROS augmentation使得模型适应various object sizes。因此，这两个方法在有着明显object size-related domain gaps的N-K和W-K的tasks中表现最好，尤其是在Car物体上。但在K-N的task中，object size的不存在显著差异，然而density-related domain gaps占据主导。在K-N的task中，我们的方法outperform SN和ST3D++。

{\noindent \textbf{Comparison with UDA methods.}} The UDA methods leverage reachable unlabeled data and relevant knowledge on the target domain to improve the detection of target domain data. SN aims to normalize object sizes to the target domain, and ST3D++ utilizes ROS augmentation to make the model adaptable to various object sizes. Hence, as shown in Table~\ref{tab: main_comparision_DG_UDA}, these two methods perform best on NuScenes $\rightarrow$ KITTI and Waymo $\rightarrow$ KITTI with noticeable domain gaps related to object size, especially for ``Car'' objects. However, on Waymo $\rightarrow$ NuScenes where there is no significant difference in object size, density-related domain gaps dominate. For Waymo $\rightarrow$ NuScenes, our method outperforms SN and ST3D++ and reaches the best performance. 

\subsection{Ablation Study}
\label{sec: alation_study}

In this section, we conduct extensive ablation experiments to investigate the individual components of our
proposed DG method. All ablation studies are conducted on NuScenes $\rightarrow$ KITTI and Waymo $\rightarrow$ NuScenes and use the VoxelRCNN as the basic detection model. More detailed results are shown in Section~\ref{sec: suppl_ablation_study} in the supplementary.  

\noindent \textbf{Component ablation.} As demonstrated
in Table~\ref{tab: component_comparison}, we investigate the effectiveness of our individual components. Compared with (a) the detection model only trained with source domain data, (b) applying PDDA augmentation during source training brings significant improvement. By means of (c) multi-task learning, the detection rate has a slight boost. In the end, (d) the adoption of test-time adaption further improves the performance.

\begin{table}[]
\centering
\caption{Component ablations in mAP(\%). \textbf{Source} represents the conventional training procedure. \textbf{PDDA}, \textbf{MTL} and \textbf{TTA} represent our data augmentation, multi-task learning, and test-time adaptation, respectively.}
\label{tab: component_comparison}
\resizebox{0.9\linewidth}{!}{%
\begin{tabular}{@{}c|cccc|c|c@{}}
\toprule
& \textbf{Source} & \textbf{PDDA} & \textbf{MTL} & \textbf{TTA} & \textbf{N$\rightarrow$K} & \textbf{W$\rightarrow$N} \\ \midrule
(a) & \checkmark &  &  &  & 35.34/18.5 & 14.16/9.36 \\
(b) &\checkmark & \checkmark &  &  & 41.07/24.08 & 16.83/11.04 \\
(c) &\checkmark & \checkmark & \checkmark &  & 41.35/24.12 & 17.09/11.17 \\ \midrule
(d) &\checkmark & \checkmark & \checkmark & \checkmark & \textbf{42.17}/\textbf{25.15} & \textbf{17.22}/\textbf{11.26} \\ \bottomrule
\end{tabular}%
}
\end{table}

\noindent \textbf{Comparison on data augmentation.} Besides the DG augmentation methods (\ie, PA-DA~\cite{choi2021part} and 3D-VF~\cite{lehner20223d}), we also include the weather-simulating augmentation methods Rain ~\cite{kilic2021lidar} and Fog~\cite{hahner2021fog} which simulate the adverse rain and fog in point clouds. As shown in Table~\ref{tab: augmentation_comparison}, our PDDA method outperforms all compared DG augmentation methods. %5.33\%/4.10\% on NuScenes $\rightarrow$ KITTI and 10.65\%/12.00\% on Waymo $\rightarrow$ NuScenes.  
In this ablation, we also include UDA augmentation methods that require target-domain information: SN~\cite{wang2020train}, RBRS~\cite{hu2023density}, and ROS~\cite{yang2021st3d++, yang2021st3d}. % as comparison baselines. The former masks or interpolates beam layers to increase the density diversity of inputs and the latter scales object sizes to mitigate the domain shift of object size statistics. 
Particularly, for density-related domain shifts, RBRS~\cite{hu2023density} employs random upsampling on the low-to-high-density NuScenes $\rightarrow$ KITTI and downsampling on the high-to-low-density Waymo $\rightarrow$ NuScenes. For comparison, without relying on target domain information, our PDDA outperforms RBRS on NuScenes $\rightarrow$ KITTI and closely approximates RBRS on Waymo $\rightarrow$ NuScenes (with a slight 0.5\%/0.16\% lag), which indicates the superiority of PDDA's capturing the physical-aware characteristics of real-world beam layer scanning.

\begin{table}[]
\centering
\caption{Data augmentation ablations in mAP(\%). The bold \textbf{values} represent the best performance in DG tasks and the underlined \underline{values} for the best performance in DG + UDA tasks.}
\label{tab: augmentation_comparison}
\resizebox{0.67\linewidth}{!}{%
\begin{tabular}{@{}cc|c|c@{}}
\toprule
\multicolumn{1}{c}{\textbf{Tasks}} & \multicolumn{1}{c|}{\textbf{Methods}} & \textbf{N$\rightarrow$K} & \textbf{W$\rightarrow$N} \\ \midrule
\multirow{6}{*}{\textbf{DG}} & Source-only & 35.34/18.50 & 14.16/9.36 \\  
 & PA-DA\cite{choi2021part} & 34.16/21.10 & 13.70/8.97 \\
 & 3D-VF\cite{lehner20223d} & 37.45/23.13 & 13.82/8.97 \\
 & Fog\cite{hahner2021fog} & 38.99/23.01 & 14.59/9.43 \\
 & Rain\cite{kilic2021lidar} & 37.04/23.12 & 15.21/9.86 \\
 & \textbf{PDDA} & \underline{\textbf{41.07}}/\textbf{24.08} & \textbf{16.83}/\textbf{11.04} \\ \midrule
\multirow{3}{*}{\textbf{UDA}} & RBRS\cite{hu2023density} & 39.21/20.82 & \underline{17.33}/\underline{11.20} \\
 & SN\cite{wang2020train} & 34.52/27.01 & 14.25/9.96 \\
 & ROS\cite{yang2021st3d++} & 38.18/\underline{28.29} & 13.67/8.86 \\ \bottomrule
\end{tabular}%
}
\end{table}

\noindent \textbf{Comparison on test-time training.} To further explore the performance of our proposed test-time training, following the Point-MATE~\cite{mirza2023mate}, we adopt the masking strategy based on K nearest neighbors (KNN) with the masking rate of 90\% during self-supervised training and test-time training (refer to implementation details in Section~\ref{sec: suppl_experiment_implimentation} in the supplementary). As shown in Table~\ref{tab: compare_TTT}, while the KNN-masking-based test-time training reaches approximate accuracy to our density-downsampling-based method (with a slight lag), the computation of farthest point sampling and KNN clustering on all points cause the severe latency. In contrast, our density-sampling operations on points perform more accurately and more efficiently on object detection.

\begin{table}[]
\centering
\caption{Comparison on test-time training. KNN-TTT stands for applying the KNN-based masking strategy during self-supervised training and test-time training, instead of density-downsampling as in our proposed method. FPS stands for the number of frames processed by the detection model per second.}
\label{tab: compare_TTT}
\resizebox{0.98\linewidth}{!}{%
\begin{tabular}{@{}c|cc|cc@{}}
\toprule
\multirow{2}{*}{Methods} & \multicolumn{2}{c|}{N$\rightarrow$K} & \multicolumn{2}{c}{W$\rightarrow$N} \\  \cmidrule(l){2-5}
 & mAP (\%) & \begin{tabular}[c|]{@{}c@{}}computation \\(FPS)\end{tabular} & mAP (\%) & \begin{tabular}[c]{@{}c@{}}computation \\(FPS)\end{tabular} \\ \midrule
 KNN-TTT\cite{mirza2023mate} & 40.51/24.92 & 0.52 & 16.67/11.03 & 0.41 \\ 
 \textbf{Our TTT}& \textbf{42.17}/\textbf{25.15} & \textbf{4.23} & \textbf{17.22}/\textbf{11.26} &  \textbf{3.52} \\ \bottomrule
\end{tabular}%
}
\end{table}

\begin{figure}[t]
  \centering
   \includegraphics[width=\linewidth]{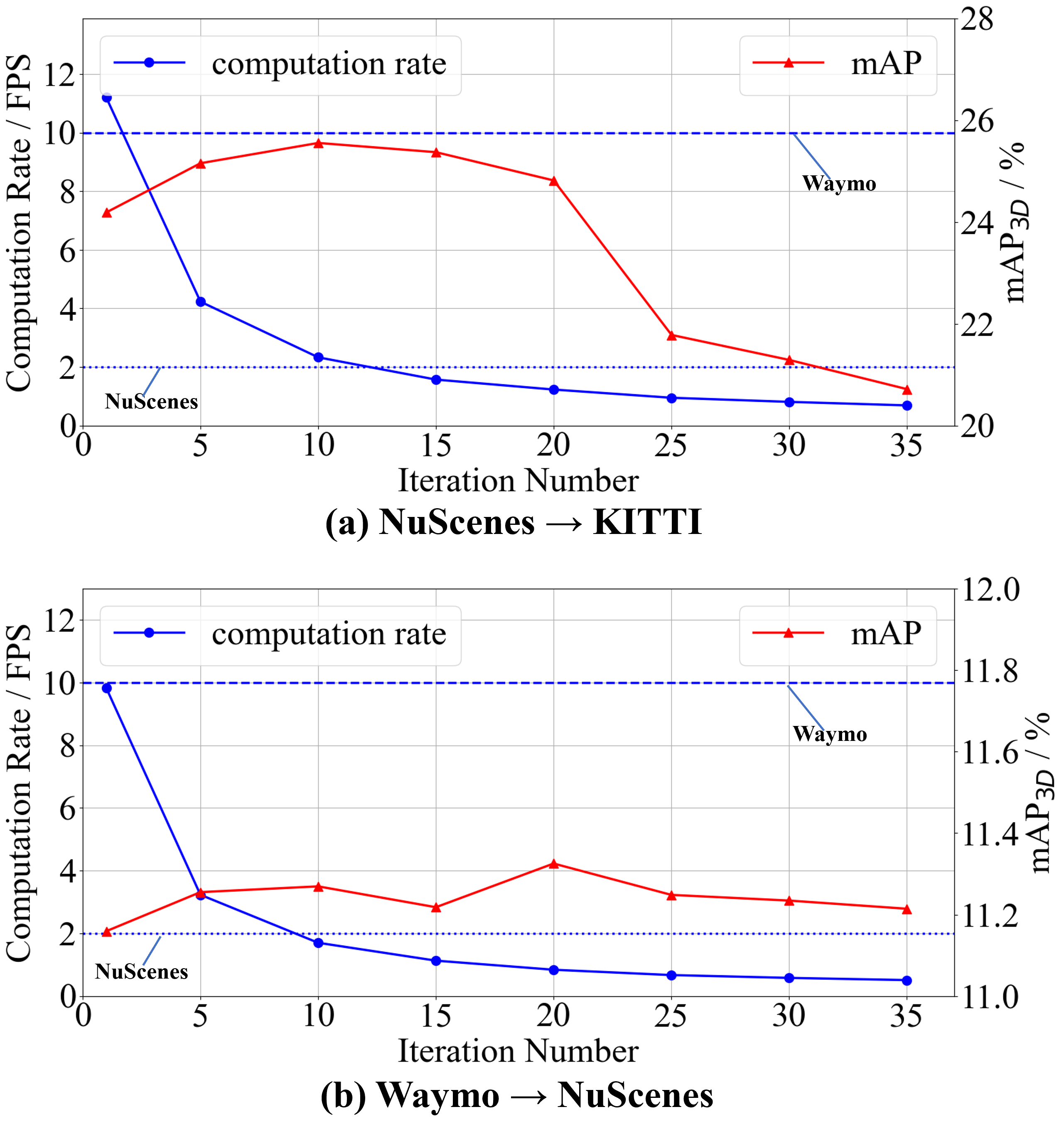}
   \caption{Computation efficiency on (a) NuScenes $\rightarrow$ KITTI and (b) Waymo $\rightarrow$ NuScenes. We indicate Waymo's frame rate of 10 FPS and NuScenes's keyframe rate of 2 FPS by dash lines.}
   \label{fig: compute_efficiency}
\end{figure}

\noindent \textbf{Computation efficiency.} We explore the computation efficiency \wrt processing of a single GeForce RTX-3090 GPU. As shown in Figure~\ref{fig: compute_efficiency}, the optimal $N_{iter}$ reaching the best performance are 10 for NuScenes $\rightarrow$ KITTI and 20 for Waymo $\rightarrow$ NuScenes. After that, the detection performance gets worse due to the model overfitting to the auxiliary task. Given such $N_{iter}$ settings, the real-time processing requires parallel computation with multiple GPUs. For the single-GPU setting, the computation speed with $N_{iter}=5$ still overpasses the 2 FPS labeled keyframe rate of NuScenes. By a minor performance penalty, the computation speed with $N_{iter}=1$ meets the 10 FPS real-time running requirement of Waymo, and the accuracy (41.41\%/24.19\% on NuScenes $\rightarrow$ KITTI and 17.00\%/11.16\%) still overpass other DG methods in Table~\ref{tab: main_comparision_DG_UDA}.

\subsection{Limitation} 
As shown in Table~\ref{tab: main_comparision_DG_UDA}, our proposed method improves the model's generalizability on density-variation-related settings, e.g. NuScenes $\rightarrow$ KITTI, Waymo $\rightarrow$ NuScenes, and ``Car'' detection on Waymo $\rightarrow$ KITTI where object size bias exists. However, regarding tasks with no significant domain shifts, such as ``Pedestrian'' and ``Cyclist'' detection on Waymo $\rightarrow$ KITTI, our method brings no or minor improvement in detection accuracy, which is a topic we plan to solve in the future.

\section{Conclusion}
Point-cloud-based 3D object detection suffers from performance degradation when encountering data with unexplored domain gaps. To tackle this problem, we proposed the domain generalization method to improve the model's generalizability. We first designed a physical-aware density-resampling data augmentation to mitigate the performance loss stemming from diverse point densities. 
From the learning methodology viewpoint, we introduced a multi-task learning solution, incorporating self-supervised 3D scene restoration into the object detection task. 
Beyond the model optimization benefit in source-domain training, the self-supervised restoration task is also used for the test-time update of the encoder for feature extraction.
As the first test-time adaptation solution on domain generalization of 3D point-cloud-based object detection, our method significantly improves the detection model's performance to unseen target domains. 
{
    \small
    \bibliographystyle{ieeenat_fullname}
    \bibliography{main}

\begin{thebibliography}{54}
\providecommand{\natexlab}[1]{#1}
\providecommand{\url}[1]{\texttt{#1}}
\expandafter\ifx\csname urlstyle\endcsname\relax
  \providecommand{\doi}[1]{doi: #1}\else
  \providecommand{\doi}{doi: \begingroup \urlstyle{rm}\Url}\fi

\bibitem[Arnold et~al.(2019)Arnold, Al-Jarrah, Dianati, Fallah, Oxtoby, and Mouzakitis]{arnold2019survey}
Eduardo Arnold, Omar~Y Al-Jarrah, Mehrdad Dianati, Saber Fallah, David Oxtoby, and Alex Mouzakitis.
\newblock A survey on 3d object detection methods for autonomous driving applications.
\newblock \emph{IEEE Transactions on Intelligent Transportation Systems}, 20\penalty0 (10):\penalty0 3782--3795, 2019.

\bibitem[Balaji et~al.(2018)Balaji, Sankaranarayanan, and Chellappa]{balaji2018metareg}
Yogesh Balaji, Swami Sankaranarayanan, and Rama Chellappa.
\newblock Metareg: Towards domain generalization using meta-regularization.
\newblock \emph{Advances in neural information processing systems}, 31, 2018.

\bibitem[Boyce et~al.(1989)Boyce, Pollatsek, and Rayner]{boyce1989effect}
Susan~J Boyce, Alexander Pollatsek, and Keith Rayner.
\newblock Effect of background information on object identification.
\newblock \emph{Journal of Experimental Psychology: Human Perception and Performance}, 15\penalty0 (3):\penalty0 556, 1989.

\bibitem[Caesar et~al.(2020)Caesar, Bankiti, Lang, Vora, Liong, Xu, Krishnan, Pan, Baldan, and Beijbom]{caesar2020nuscenes}
Holger Caesar, Varun Bankiti, Alex~H Lang, Sourabh Vora, Venice~Erin Liong, Qiang Xu, Anush Krishnan, Yu Pan, Giancarlo Baldan, and Oscar Beijbom.
\newblock nuscenes: A multimodal dataset for autonomous driving.
\newblock In \emph{Proceedings of the IEEE/CVF conference on computer vision and pattern recognition}, pages 11621--11631, 2020.

\bibitem[Chen et~al.(2022{\natexlab{a}})Chen, Wang, Darrell, and Ebrahimi]{chen2022contrastive}
Dian Chen, Dequan Wang, Trevor Darrell, and Sayna Ebrahimi.
\newblock Contrastive test-time adaptation.
\newblock In \emph{Proceedings of the IEEE/CVF Conference on Computer Vision and Pattern Recognition}, pages 295--305, 2022{\natexlab{a}}.

\bibitem[Chen et~al.(2023)Chen, Zhang, Song, Shan, and Liu]{chen2023improved}
Liang Chen, Yong Zhang, Yibing Song, Ying Shan, and Lingqiao Liu.
\newblock Improved test-time adaptation for domain generalization.
\newblock In \emph{Proceedings of the IEEE/CVF Conference on Computer Vision and Pattern Recognition}, pages 24172--24182, 2023.

\bibitem[Chen et~al.(2022{\natexlab{b}})Chen, Lin, Yang, Xie, Pu, and Zhuang]{chen2022self}
Weijie Chen, Luojun Lin, Shicai Yang, Di Xie, Shiliang Pu, and Yueting Zhuang.
\newblock Self-supervised noisy label learning for source-free unsupervised domain adaptation.
\newblock In \emph{2022 IEEE/RSJ International Conference on Intelligent Robots and Systems (IROS)}, pages 10185--10192. IEEE, 2022{\natexlab{b}}.

\bibitem[Choi et~al.(2021)Choi, Song, and Kwak]{choi2021part}
Jaeseok Choi, Yeji Song, and Nojun Kwak.
\newblock Part-aware data augmentation for 3d object detection in point cloud.
\newblock In \emph{2021 IEEE/RSJ International Conference on Intelligent Robots and Systems (IROS)}, pages 3391--3397. IEEE, 2021.

\bibitem[Deng et~al.(2021)Deng, Shi, Li, Zhou, Zhang, and Li]{deng2021voxel}
Jiajun Deng, Shaoshuai Shi, Peiwei Li, Wengang Zhou, Yanyong Zhang, and Houqiang Li.
\newblock Voxel r-cnn: Towards high performance voxel-based 3d object detection.
\newblock In \emph{Proceedings of the AAAI Conference on Artificial Intelligence}, pages 1201--1209, 2021.

\bibitem[Geiger et~al.(2013)Geiger, Lenz, Stiller, and Urtasun]{geiger2013vision}
Andreas Geiger, Philip Lenz, Christoph Stiller, and Raquel Urtasun.
\newblock Vision meets robotics: The kitti dataset.
\newblock \emph{The International Journal of Robotics Research}, 32\penalty0 (11):\penalty0 1231--1237, 2013.

\bibitem[Hahner et~al.(2021)Hahner, Sakaridis, Dai, and Van~Gool]{hahner2021fog}
Martin Hahner, Christos Sakaridis, Dengxin Dai, and Luc Van~Gool.
\newblock Fog simulation on real lidar point clouds for 3d object detection in adverse weather.
\newblock In \emph{Proceedings of the IEEE/CVF International Conference on Computer Vision}, pages 15283--15292, 2021.

\bibitem[Hahner et~al.(2022)Hahner, Sakaridis, Bijelic, Heide, Yu, Dai, and Van~Gool]{hahner2022lidar}
Martin Hahner, Christos Sakaridis, Mario Bijelic, Felix Heide, Fisher Yu, Dengxin Dai, and Luc Van~Gool.
\newblock Lidar snowfall simulation for robust 3d object detection.
\newblock In \emph{Proceedings of the IEEE/CVF Conference on Computer Vision and Pattern Recognition}, pages 16364--16374, 2022.

\bibitem[Hammer et~al.(2018)Hammer, Hebel, Laurenzis, and Arens]{hammer2018lidar}
Marcus Hammer, Marcus Hebel, Martin Laurenzis, and Michael Arens.
\newblock Lidar-based detection and tracking of small uavs.
\newblock In \emph{Emerging Imaging and Sensing Technologies for Security and Defence III; and Unmanned Sensors, Systems, and Countermeasures}, pages 177--185. SPIE, 2018.

\bibitem[Hu et~al.(2022)Hu, Kuai, and Waslander]{hu2022point}
Jordan~SK Hu, Tianshu Kuai, and Steven~L Waslander.
\newblock Point density-aware voxels for lidar 3d object detection.
\newblock In \emph{Proceedings of the IEEE/CVF Conference on Computer Vision and Pattern Recognition}, pages 8469--8478, 2022.

\bibitem[Hu et~al.(2023)Hu, Liu, and Hu]{hu2023density}
Qianjiang Hu, Daizong Liu, and Wei Hu.
\newblock Density-insensitive unsupervised domain adaption on 3d object detection.
\newblock In \emph{Proceedings of the IEEE/CVF Conference on Computer Vision and Pattern Recognition}, pages 17556--17566, 2023.

\bibitem[Iwasawa and Matsuo(2021)]{iwasawa2021test}
Yusuke Iwasawa and Yutaka Matsuo.
\newblock Test-time classifier adjustment module for model-agnostic domain generalization.
\newblock \emph{Advances in Neural Information Processing Systems}, 34:\penalty0 2427--2440, 2021.

\bibitem[Kilic et~al.(2021)Kilic, Hegde, Sindagi, Cooper, Foster, and Patel]{kilic2021lidar}
Velat Kilic, Deepti Hegde, Vishwanath Sindagi, A~Brinton Cooper, Mark~A Foster, and Vishal~M Patel.
\newblock Lidar light scattering augmentation (lisa): Physics-based simulation of adverse weather conditions for 3d object detection.
\newblock \emph{arXiv preprint arXiv:2107.07004}, 2021.

\bibitem[Lang et~al.(2019)Lang, Vora, Caesar, Zhou, Yang, and Beijbom]{lang2019pointpillars}
Alex~H Lang, Sourabh Vora, Holger Caesar, Lubing Zhou, Jiong Yang, and Oscar Beijbom.
\newblock Pointpillars: Fast encoders for object detection from point clouds.
\newblock In \emph{Proceedings of the IEEE/CVF conference on computer vision and pattern recognition}, pages 12697--12705, 2019.

\bibitem[Lehner et~al.(2022)Lehner, Gasperini, Marcos-Ramiro, Schmidt, Mahani, Navab, Busam, and Tombari]{lehner20223d}
Alexander Lehner, Stefano Gasperini, Alvaro Marcos-Ramiro, Michael Schmidt, Mohammad-Ali~Nikouei Mahani, Nassir Navab, Benjamin Busam, and Federico Tombari.
\newblock 3d-vfield: Adversarial augmentation of point clouds for domain generalization in 3d object detection.
\newblock In \emph{Proceedings of the IEEE/CVF Conference on Computer Vision and Pattern Recognition}, pages 17295--17304, 2022.

\bibitem[Li et~al.(2018)Li, Yang, Song, and Hospedales]{li2018learning}
Da Li, Yongxin Yang, Yi-Zhe Song, and Timothy Hospedales.
\newblock Learning to generalize: Meta-learning for domain generalization.
\newblock In \emph{Proceedings of the AAAI conference on artificial intelligence}, 2018.

\bibitem[Liang et~al.(2020)Liang, Hu, and Feng]{liang2020we}
Jian Liang, Dapeng Hu, and Jiashi Feng.
\newblock Do we really need to access the source data? source hypothesis transfer for unsupervised domain adaptation.
\newblock In \emph{International conference on machine learning}, pages 6028--6039. PMLR, 2020.

\bibitem[Liang et~al.(2023)Liang, He, and Tan]{liang2023comprehensive}
Jian Liang, Ran He, and Tieniu Tan.
\newblock A comprehensive survey on test-time adaptation under distribution shifts.
\newblock \emph{arXiv preprint arXiv:2303.15361}, 2023.

\bibitem[Liu et~al.(2022)Liu, Chen, Dou, and Heng]{liu2022single}
Quande Liu, Cheng Chen, Qi Dou, and Pheng-Ann Heng.
\newblock Single-domain generalization in medical image segmentation via test-time adaptation from shape dictionary.
\newblock In \emph{Proceedings of the AAAI Conference on Artificial Intelligence}, pages 1756--1764, 2022.

\bibitem[Mao et~al.(2021)Mao, Xue, Niu, Bai, Feng, Liang, Xu, and Xu]{mao2021voxel}
Jiageng Mao, Yujing Xue, Minzhe Niu, Haoyue Bai, Jiashi Feng, Xiaodan Liang, Hang Xu, and Chunjing Xu.
\newblock Voxel transformer for 3d object detection.
\newblock In \emph{Proceedings of the IEEE/CVF International Conference on Computer Vision}, pages 3164--3173, 2021.

\bibitem[Mirza et~al.(2023)Mirza, Shin, Lin, Schriebl, Sun, Choe, Kozinski, Possegger, Kweon, Yoon, et~al.]{mirza2023mate}
M~Jehanzeb Mirza, Inkyu Shin, Wei Lin, Andreas Schriebl, Kunyang Sun, Jaesung Choe, Mateusz Kozinski, Horst Possegger, In~So Kweon, Kuk-Jin Yoon, et~al.
\newblock Mate: Masked autoencoders are online 3d test-time learners.
\newblock In \emph{Proceedings of the IEEE/CVF International Conference on Computer Vision}, pages 16709--16718, 2023.

\bibitem[Motiian et~al.(2017)Motiian, Piccirilli, Adjeroh, and Doretto]{motiian2017unified}
Saeid Motiian, Marco Piccirilli, Donald~A Adjeroh, and Gianfranco Doretto.
\newblock Unified deep supervised domain adaptation and generalization.
\newblock In \emph{Proceedings of the IEEE international conference on computer vision}, pages 5715--5725, 2017.

\bibitem[Muandet et~al.(2013)Muandet, Balduzzi, and Sch{\"o}lkopf]{muandet2013domain}
Krikamol Muandet, David Balduzzi, and Bernhard Sch{\"o}lkopf.
\newblock Domain generalization via invariant feature representation.
\newblock In \emph{International conference on machine learning}, pages 10--18. PMLR, 2013.

\bibitem[Ouyang et~al.(2022)Ouyang, Chen, Li, Li, Qin, Bai, and Rueckert]{ouyang2022causality}
Cheng Ouyang, Chen Chen, Surui Li, Zeju Li, Chen Qin, Wenjia Bai, and Daniel Rueckert.
\newblock Causality-inspired single-source domain generalization for medical image segmentation.
\newblock \emph{IEEE Transactions on Medical Imaging}, 42\penalty0 (4):\penalty0 1095--1106, 2022.

\bibitem[Qi et~al.(2017)Qi, Yi, Su, and Guibas]{qi2017pointnet++}
Charles~Ruizhongtai Qi, Li Yi, Hao Su, and Leonidas~J Guibas.
\newblock Pointnet++: Deep hierarchical feature learning on point sets in a metric space.
\newblock \emph{Advances in neural information processing systems}, 30, 2017.

\bibitem[Qian et~al.(2022)Qian, Lai, and Li]{qian20223d}
Rui Qian, Xin Lai, and Xirong Li.
\newblock 3d object detection for autonomous driving: A survey.
\newblock \emph{Pattern Recognition}, 130:\penalty0 108796, 2022.

\bibitem[Shi et~al.(2019)Shi, Wang, and Li]{shi2019pointrcnn}
Shaoshuai Shi, Xiaogang Wang, and Hongsheng Li.
\newblock Pointrcnn: 3d object proposal generation and detection from point cloud.
\newblock In \emph{Proceedings of the IEEE/CVF conference on computer vision and pattern recognition}, pages 770--779, 2019.

\bibitem[Shi et~al.(2020)Shi, Guo, Jiang, Wang, Shi, Wang, and Li]{shi2020pv}
Shaoshuai Shi, Chaoxu Guo, Li Jiang, Zhe Wang, Jianping Shi, Xiaogang Wang, and Hongsheng Li.
\newblock Pv-rcnn: Point-voxel feature set abstraction for 3d object detection.
\newblock In \emph{Proceedings of the IEEE/CVF conference on computer vision and pattern recognition}, pages 10529--10538, 2020.

\bibitem[Shi et~al.(2023)Shi, Jiang, Deng, Wang, Guo, Shi, Wang, and Li]{shi2023pv}
Shaoshuai Shi, Li Jiang, Jiajun Deng, Zhe Wang, Chaoxu Guo, Jianping Shi, Xiaogang Wang, and Hongsheng Li.
\newblock Pv-rcnn++: Point-voxel feature set abstraction with local vector representation for 3d object detection.
\newblock \emph{International Journal of Computer Vision}, 131\penalty0 (2):\penalty0 531--551, 2023.

\bibitem[Shi and Rajkumar(2020)]{shi2020point}
Weijing Shi and Raj Rajkumar.
\newblock Point-gnn: Graph neural network for 3d object detection in a point cloud.
\newblock In \emph{Proceedings of the IEEE/CVF conference on computer vision and pattern recognition}, pages 1711--1719, 2020.

\bibitem[Soum-Fontez et~al.(2023)Soum-Fontez, Deschaud, and Goulette]{soum2023mdt3d}
Louis Soum-Fontez, Jean-Emmanuel Deschaud, and Fran{\c{c}}ois Goulette.
\newblock Mdt3d: Multi-dataset training for lidar 3d object detection generalization.
\newblock \emph{arXiv preprint arXiv:2308.01000}, 2023.

\bibitem[Sun et~al.(2020)Sun, Kretzschmar, Dotiwalla, Chouard, Patnaik, Tsui, Guo, Zhou, Chai, Caine, et~al.]{sun2020scalability}
Pei Sun, Henrik Kretzschmar, Xerxes Dotiwalla, Aurelien Chouard, Vijaysai Patnaik, Paul Tsui, James Guo, Yin Zhou, Yuning Chai, Benjamin Caine, et~al.
\newblock Scalability in perception for autonomous driving: Waymo open dataset.
\newblock In \emph{Proceedings of the IEEE/CVF conference on computer vision and pattern recognition}, pages 2446--2454, 2020.

\bibitem[Team(2020)]{openpcdet2020}
OpenPCDet~Development Team.
\newblock Openpcdet: An open-source toolbox for 3d object detection from point clouds.
\newblock \url{https://github.com/open-mmlab/OpenPCDet}, 2020.

\bibitem[Vidit et~al.(2023)Vidit, Engilberge, and Salzmann]{vidit2023clip}
Vidit Vidit, Martin Engilberge, and Mathieu Salzmann.
\newblock Clip the gap: A single domain generalization approach for object detection.
\newblock In \emph{Proceedings of the IEEE/CVF Conference on Computer Vision and Pattern Recognition}, pages 3219--3229, 2023.

\bibitem[Volpi and Murino(2019)]{volpi2019addressing}
Riccardo Volpi and Vittorio Murino.
\newblock Addressing model vulnerability to distributional shifts over image transformation sets.
\newblock In \emph{Proceedings of the IEEE/CVF International Conference on Computer Vision}, pages 7980--7989, 2019.

\bibitem[Volpi et~al.(2018)Volpi, Namkoong, Sener, Duchi, Murino, and Savarese]{volpi2018generalizing}
Riccardo Volpi, Hongseok Namkoong, Ozan Sener, John~C Duchi, Vittorio Murino, and Silvio Savarese.
\newblock Generalizing to unseen domains via adversarial data augmentation.
\newblock \emph{Advances in neural information processing systems}, 31, 2018.

\bibitem[Wang et~al.(2020{\natexlab{a}})Wang, Shelhamer, Liu, Olshausen, and Darrell]{wang2020tent}
Dequan Wang, Evan Shelhamer, Shaoteng Liu, Bruno Olshausen, and Trevor Darrell.
\newblock Tent: Fully test-time adaptation by entropy minimization.
\newblock \emph{arXiv preprint arXiv:2006.10726}, 2020{\natexlab{a}}.

\bibitem[Wang et~al.(2020{\natexlab{b}})Wang, Chen, You, Li, Hariharan, Campbell, Weinberger, and Chao]{wang2020train}
Yan Wang, Xiangyu Chen, Yurong You, Li~Erran Li, Bharath Hariharan, Mark Campbell, Kilian~Q Weinberger, and Wei-Lun Chao.
\newblock Train in germany, test in the usa: Making 3d object detectors generalize.
\newblock In \emph{Proceedings of the IEEE/CVF Conference on Computer Vision and Pattern Recognition}, pages 11713--11723, 2020{\natexlab{b}}.

\bibitem[Wang et~al.(2021)Wang, Luo, Qiu, Huang, and Baktashmotlagh]{wang2021learning}
Zijian Wang, Yadan Luo, Ruihong Qiu, Zi Huang, and Mahsa Baktashmotlagh.
\newblock Learning to diversify for single domain generalization.
\newblock In \emph{Proceedings of the IEEE/CVF International Conference on Computer Vision}, pages 834--843, 2021.

\bibitem[Wu and Deng(2022)]{wu2022single}
Aming Wu and Cheng Deng.
\newblock Single-domain generalized object detection in urban scene via cyclic-disentangled self-distillation.
\newblock In \emph{Proceedings of the IEEE/CVF Conference on computer vision and pattern recognition}, pages 847--856, 2022.

\bibitem[Wu et~al.(2023)Wu, Cao, Liu, Chen, and Ren]{wu2023towards}
Guile Wu, Tongtong Cao, Bingbing Liu, Xingxin Chen, and Yuan Ren.
\newblock Towards universal lidar-based 3d object detection by multi-domain knowledge transfer.
\newblock In \emph{Proceedings of the IEEE/CVF International Conference on Computer Vision}, pages 8669--8678, 2023.

\bibitem[Xu et~al.(2022)Xu, Zhong, and Neumann]{xu2022behind}
Qiangeng Xu, Yiqi Zhong, and Ulrich Neumann.
\newblock Behind the curtain: Learning occluded shapes for 3d object detection.
\newblock In \emph{Proceedings of the AAAI Conference on Artificial Intelligence}, pages 2893--2901, 2022.

\bibitem[Yan et~al.(2018)Yan, Mao, and Li]{yan2018second}
Yan Yan, Yuxing Mao, and Bo Li.
\newblock Second: Sparsely embedded convolutional detection.
\newblock \emph{Sensors}, 18\penalty0 (10):\penalty0 3337, 2018.

\bibitem[Yang et~al.(2021{\natexlab{a}})Yang, Shi, Wang, Li, and Qi]{yang2021st3d}
Jihan Yang, Shaoshuai Shi, Zhe Wang, Hongsheng Li, and Xiaojuan Qi.
\newblock St3d: Self-training for unsupervised domain adaptation on 3d object detection.
\newblock In \emph{Proceedings of the IEEE/CVF conference on computer vision and pattern recognition}, pages 10368--10378, 2021{\natexlab{a}}.

\bibitem[Yang et~al.(2021{\natexlab{b}})Yang, Shi, Wang, Li, and Qi]{yang2021st3d++}
Jihan Yang, Shaoshuai Shi, Zhe Wang, Hongsheng Li, and Xiaojuan Qi.
\newblock St3d++: denoised self-training for unsupervised domain adaptation on 3d object detection.
\newblock \emph{arXiv preprint arXiv:2108.06682}, 2021{\natexlab{b}}.

\bibitem[Yang et~al.(2018)Yang, Yan, Zhang, Yu, Shi, Mou, Kalra, Zhang, Sun, and Wang]{yang2018low}
Qingsong Yang, Pingkun Yan, Yanbo Zhang, Hengyong Yu, Yongyi Shi, Xuanqin Mou, Mannudeep~K Kalra, Yi Zhang, Ling Sun, and Ge Wang.
\newblock Low-dose ct image denoising using a generative adversarial network with wasserstein distance and perceptual loss.
\newblock \emph{IEEE transactions on medical imaging}, 37\penalty0 (6):\penalty0 1348--1357, 2018.

\bibitem[Ye et~al.(2020)Ye, Chen, Zhang, Hao, and Zhang]{ye2020sarpnet}
Yangyang Ye, Houjin Chen, Chi Zhang, Xiaoli Hao, and Zhaoxiang Zhang.
\newblock Sarpnet: Shape attention regional proposal network for lidar-based 3d object detection.
\newblock \emph{Neurocomputing}, 379:\penalty0 53--63, 2020.

\bibitem[Zhang et~al.(2023)Zhang, Yuan, Shi, Chen, Li, and Qiao]{zhang2023uni3d}
Bo Zhang, Jiakang Yuan, Botian Shi, Tao Chen, Yikang Li, and Yu Qiao.
\newblock Uni3d: A unified baseline for multi-dataset 3d object detection.
\newblock In \emph{Proceedings of the IEEE/CVF Conference on Computer Vision and Pattern Recognition}, pages 9253--9262, 2023.

\bibitem[Zheng et~al.(2022)Zheng, Duan, Lu, Zhou, and Tian]{zheng2022hyperdet3d}
Yu Zheng, Yueqi Duan, Jiwen Lu, Jie Zhou, and Qi Tian.
\newblock Hyperdet3d: Learning a scene-conditioned 3d object detector.
\newblock In \emph{Proceedings of the IEEE/CVF Conference on Computer Vision and Pattern Recognition}, pages 5585--5594, 2022.

\bibitem[Zhou et~al.(2022)Zhou, Liu, Qiao, Xiang, and Loy]{zhou2022domain}
Kaiyang Zhou, Ziwei Liu, Yu Qiao, Tao Xiang, and Chen~Change Loy.
\newblock Domain generalization: A survey.
\newblock \emph{IEEE Transactions on Pattern Analysis and Machine Intelligence}, 2022.

\end{thebibliography}
}

\clearpage
\setcounter{page}{1}
\maketitlesupplementary

In this supplementary section, we present the visualizations of our proposed PDDA in Section~\ref{sec: suppl_PDDA} and provide more specifics for the experiment implementation and comparison results in Section~\ref{sec: suppl_experiment}.

\begin{figure}[t]
  \centering
   \includegraphics[width=\linewidth]{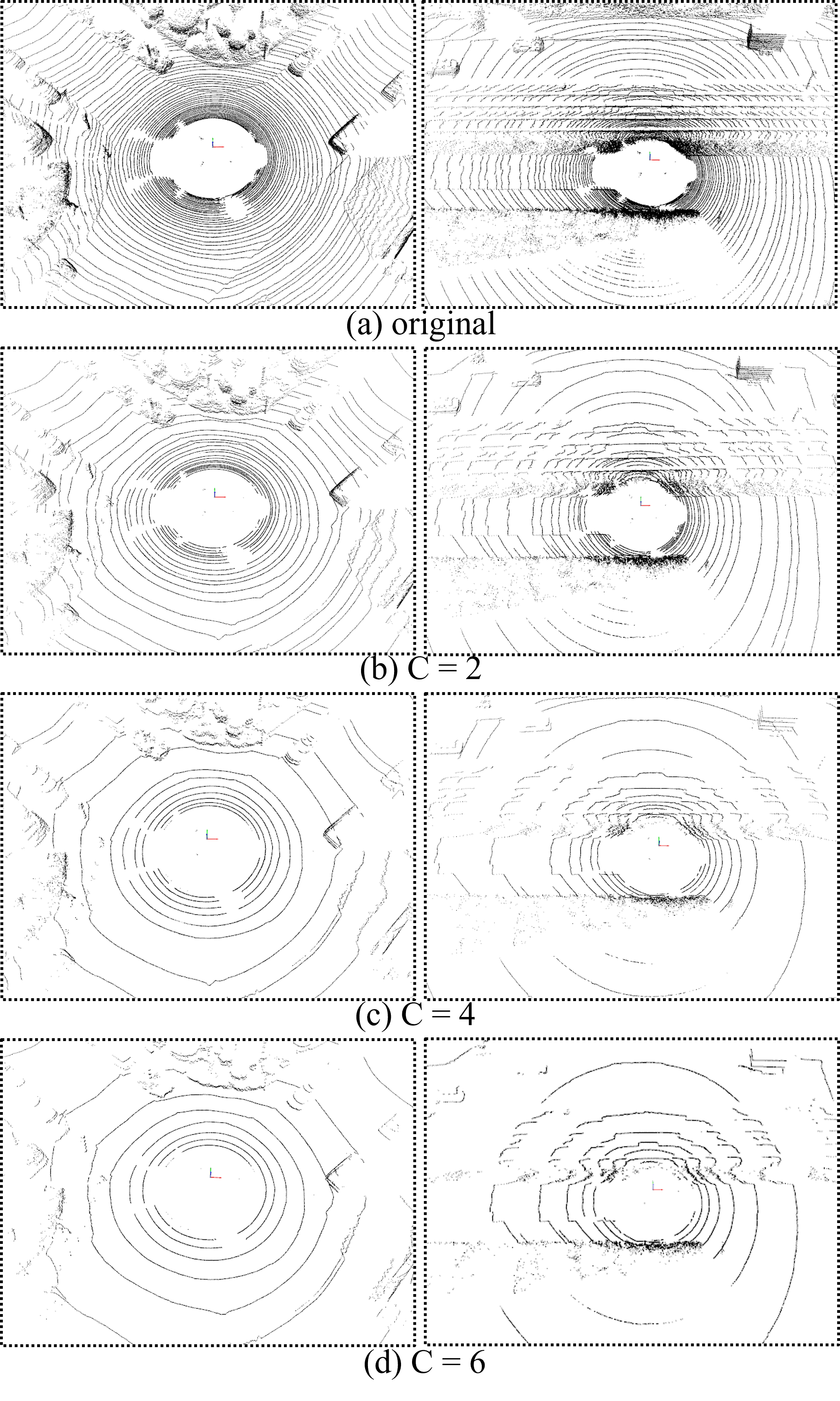}
   \caption{Visualization of density down-sampling.}
   \label{fig: suppl_vis_downsampling}
\end{figure}

\section{Physical-aware Density-resampling Data Augmentation}
\label{sec: suppl_PDDA}
\begin{figure}[t]
  \centering
   \includegraphics[width=\linewidth]{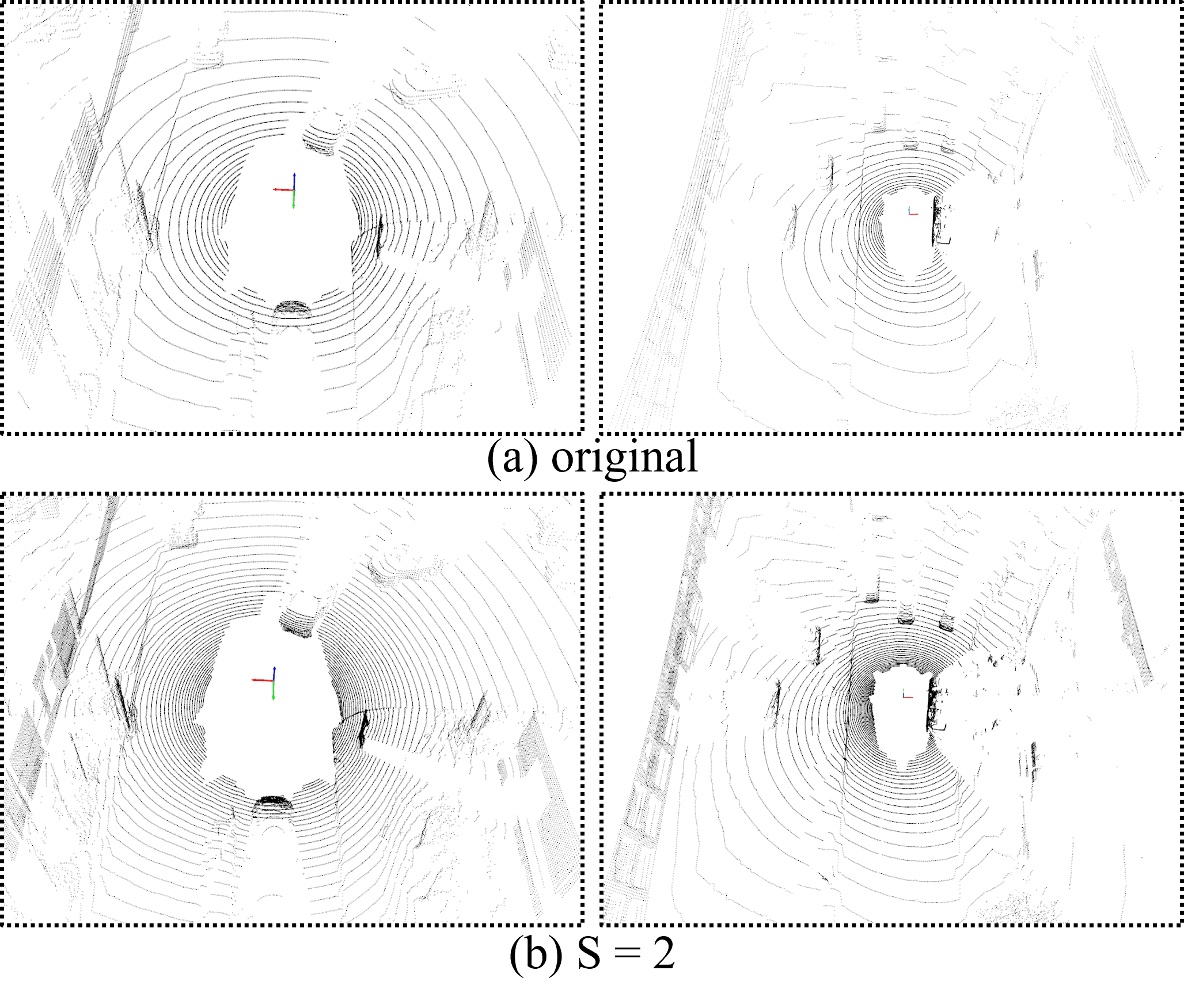}
   \caption{Visualization of density up-sampling.}
   \label{fig: suppl_vis_upsampling}
\end{figure}

In Section~\ref{sec: uni_den_aug}, we design a universal physical-aware density-resampling data augmentation (PDDA) method by simulating real-world diverse uniformly distributed beam layers of different types of sensors. For a clear visualization of PDDA's realistic point imaging, we present the density-downsampling on the 64-beam KITTI point clouds and density-upsampling on the 32-beam NuScenes point clouds. As shown in Figures~\ref{fig: suppl_vis_downsampling} and \ref{fig: suppl_vis_upsampling}, our PDDA augmentation simulates the realistic point clouds with various uniformly distributed point layers.

\begin{table}[]
\centering
\caption{Effects of the perceptual loss on detection performance}
\label{tab: Suppl_effect_pcp_loss}
\resizebox{\linewidth}{!}{%
\begin{tabular}{@{}c|cc|cc@{}}
\toprule
\multirow{2}{*}{\textbf{Losses}} & \multicolumn{2}{c|}{\textbf{N$\rightarrow$K}} & \multicolumn{2}{c}{\textbf{W$\rightarrow$N}} \\  \cmidrule(l){2-5}
 & mAP (\%) & \begin{tabular}[c|]{@{}c@{}}computation \\(FPS)\end{tabular} & mAP (\%) & \begin{tabular}[c]{@{}c@{}}computation \\(FPS)\end{tabular} \\ \midrule
 Perceptual + MSE& 42.21/25.16 & 0.89 & 17.20/11.36 & 0.78 \\ 
 MSE& 42.17/25.15 & 4.23 & 17.22/11.26 &  3.52 \\ \bottomrule
\end{tabular}%
}
\end{table}

\section{Experiment}
\label{sec: suppl_experiment} 
In this section, we present readers with more implementation specifics about the compared DG and UDA methods in Section~\ref{sec: suppl_experiment_implimentation} and show more detailed evaluation results on the detection of ``Car'', ``Pedestrian'', and ``Cyclist'' in Section~\ref{sec: suppl_ablation_study}. 

\subsection{Experiment Implementation}
\label{sec: suppl_experiment_implimentation}
Regarding compared methods shown in Sections~\ref{sec: main_experiment} and \ref{sec: alation_study}, we all apply them on a single unified VoxelRCNN model to simultaneously detect ``Car'', ``Pedestrian'', and ``Cyclist'' for the fair comparison. By following the recommended settings in the papers, we implement them for the ablation comparison study:
\begin{itemize}
    \item \textbf{3D-VF}: Due to the lack of the released code, we re-implement 3D-VF by following instructions in \cite{lehner20223d}, and apply it to objects of pedestrians and cyclists. As indicated in \cite{lehner20223d}, the augmentation by 3D-VF only randomly selects one object for each object class and perturbs its points in the point cloud. 
    \item \textbf{PA-DA}: As the recommended settings in \cite{choi2021part}, we set the parameter of augmentation as ``\textit{dropout\_p02\_swap \_p02\_mix\_p02\_sparse40\_p01\_noise10\_p01}'' and randomly augment 50\% source samples during source training.
    \item \textbf{SN}: SN~\cite{wang2020train} normalizes car sizes on the source domain with statistics of car sizes on the target domain. Given our experimental settings, we also apply it to pedestrians and cyclists.
    \item \textbf{ST3D++} and \textbf{ROS}: Follow the recommended settings and the released code in \cite{yang2021st3d++}, we apply the UDA ST3D++ and its ROS augmentation to the unified VoxelRCNN model to detect all objects of interest. 
    \item \textbf{Rain}: According to \cite{kilic2021lidar}, the severity level of rain simulation is determined by the parameter \textit{rainfall rate}. After investigation, we randomly select one \textit{rainfall rate} from the broad range $\{10, 20, 30, 40, 50\}$mm/hr to simulate the rain in the point cloud.   
    \item \textbf{Fog}: Following the recommended settings in \cite{hahner2021fog}, we randomly select one severity parameter $\alpha$ from the range $\{0, 0.005, 0.01, 0.02, 0.03, 0.06\}$ to simulate the fog in the point cloud.
    \item \textbf{RBRS}: Following the recommended settings in \cite{hu2023density}, we adopt RBRS to augment source data on different domains, specifically employing random beam upsampling on NuScenes samples and random beam downsampling on Waymo samples.
    \item \textbf{KNN-TTT}: Following \cite{mirza2023mate}, during self-supervised training and test-time adaptation, we first utilize farthest point sampling to sample 128 keypoints in the point cloud and cluster all points by KNN. Then we down-sample the points by randomly masking 90\% clusters as the settings in \cite{mirza2023mate}.
\end{itemize}

\noindent \textbf{Our proposed method.} PDDA in Section~\ref{sec: uni_den_aug} and density down-sampling operations in Sections~\ref{sec: den_down_self_train} and \ref{sec: test_time_train} work the pre-processing data augmentations to diversify the point clouds. In particular, the density down-sampling in test-time adaptation is applied as the test-time augmentation and only original point clouds before the density down-sampling are used in the target domain testing for the fair model evaluation.

Regarding the implementation of the decoder, we utilize the \textit{SparseInverseConv3d} of the Python package \textit{spconv} to conduct 8$\times$up-sampling to restore the point cloud with the original point density, considering the 8$\times$down-sampling by\textit{SparseConv3d} adopted in the encoder module of the VoxelRCNN model.

\begin{table}[t]
\centering
\caption{Data augmentation ablations in AP(\%) on NuScenes $\rightarrow$ KITTI. The bold \textbf{values} represent the best performance in DG tasks and the underlined \underline{values} for the best performance in DG + UDA tasks.}
\label{tab: Suppl_N_to_K_ablation_augment}
\resizebox{\linewidth}{!}{%
\begin{tabular}{@{}cccccc@{}}
\toprule
\multicolumn{1}{c}{\textbf{Tasks}} & \multicolumn{1}{c}{\textbf{Methods}} & \textbf{Car} & \textbf{Pedestrian} & \textbf{Cyclist} & \textbf{mAP} \\ \midrule
\multirow{6}{*}{\textbf{DG}} & Source-only & 62.69/19.02 & 22.72/18.37 & 20.61/18.13 & 35.34/18.5 \\
 & PA-DA & 65.09/\textbf{32.44} & 18.73/14.94 & 18.66/15.91 & 34.16/21.1 \\
 & 3D-VF & 65.36/29.21 & 24.85/20.87 & 22.13/19.31 & 37.45/23.13 \\
 & Fog & 68.16/29.54 & 24.55/19.24 & \textbf{24.26/20.25} & 38.99/23.01 \\
 & Rain & 66.04/30.58 & 23.81/19.56 & 21.26/19.24 & 37.04/23.12 \\
 & \textbf{PDDA} & \textbf{72.69}/31.83 & \textbf{\underline{27.1}/\underline{20.93}} & 23.41/19.48 & \textbf{\underline{41.07}/24.08} \\ \midrule
\multirow{3}{*}{\textbf{UDA}} & RBRS & 65.86/22.78 & 23.07/17.01 & \underline{28.71}/\underline{22.67} & 39.21/20.82 \\ 
 & SN & 70.5/\underline{54.78} & 19.71/15.42 & 13.36/10.83 & 34.52/27.01 \\ 
 & ROS & \underline{76.29}/53.07 & 17.41/13.34 & 20.85/18.46 & 38.18/\underline{28.29} \\ \bottomrule
\end{tabular}%
}
\end{table}

\subsection{Ablation Study}
\label{sec: suppl_ablation_study}

Table~\ref{tab: Suppl_effect_pcp_loss} shows the performance of the test-time adaptation with/without the perceptual loss. It illustrates that while additionally adopting the perceptual loss brings slight improvement during test-time parameter updating, it worsens the computing burden and causes severe computational latency.

\begin{table}[]
\centering
\caption{Data augmentation ablations in AP(\%) on Waymo $\rightarrow$ NuScenes. The bold \textbf{values} represent the best performance in DG tasks and the underlined \underline{values} for the best performance in DG + UDA tasks.}
\label{tab: Suppl_W_to_K_ablation_augment}
\resizebox{\linewidth}{!}{%
\begin{tabular}{@{}cccccc@{}}
\toprule
\multicolumn{1}{c}{\textbf{Tasks}} & \multicolumn{1}{c}{\textbf{Methods}} & \textbf{Car} & \textbf{Pedestrian} & \textbf{Cyclist} & \textbf{mAP} \\ \midrule
\multirow{6}{*}{\textbf{DG}} & Source-only & 31.20/19.13 & 10.52/8.39 & 0.75/0.55 & 14.16/9.36 \\
 & PA-DA & 29.43/18.06 & 10.84/8.43 & 0.82/0.43 & 13.70/8.97 \\
 & 3D-VF & 30.17/18.91 & 10.54/7.23 & 0.76/0.78 & 13.82/8.97 \\
 & Fog & 31.50/19.34 & 11.65/8.82 & 0.61/0.14 & 14.59/9.43 \\
 & Rain & 32.68/19.55 & 12.08/9.29 & 0.87/0.74 & 15.21/9.86 \\
 & \textbf{PDDA} & \textbf{\underline{35.67}/\underline{22.25}} & \textbf{13.79/10.08} & \textbf{\underline{1.03}/\underline{0.79}} & \textbf{16.83/11.04} \\ \midrule
\multirow{3}{*}{\textbf{UDA}} & RBRS & 35.04/21.43 & \underline{16.00}/\underline{11.50} & 0.95/0.67 & \underline{17.33}/\underline{11.20} \\ 
 & SN & 29.32/18.84 & 12.72/10.57 & 0.72/0.45 & 14.25/9.96 \\ 
 & ROS & 29.36/17.42 & 10.76/8.53 & 0.88/0.62 & 13.67/8.86 \\ \bottomrule
\end{tabular}%
}
\end{table}

Tables~\ref{tab: Suppl_N_to_K_ablation_augment} and \ref{tab: Suppl_W_to_K_ablation_augment} depict the performance of different augmentation methods on the detection of all object classes on NuScenes $\rightarrow$ KITTI and Waymo $\rightarrow$ NuScenes, respectively. As shown in Table~\ref{tab: Suppl_N_to_K_ablation_augment}, despite not achieving the best performance on all object classes among DG augmentation methods, our proposed PDDA has the best accuracy averaged on all object classes on NuScenes $\rightarrow$ KITTI. Table~\ref{tab: Suppl_W_to_K_ablation_augment} indicates the best performance of PDDA among all DG augmentation methods on the detection of all object classes and even the best performance of PDDA among all DG+UDA augmentation methods on the detection of ``Car'' and ``Cyclist'' on Waymo $\rightarrow$ NuScenes.

\begin{table*}[t]
\centering
\caption{Component ablations in AP(\%) on NuScenes $\rightarrow$ KITTI. \textbf{Source} represents the conventional training procedure. \textbf{PDDA}, \textbf{MTL} and \textbf{TTA} represent our data augmentation, multi-task learning, and test-time adaptation, respectively. The bold \textbf{values} represent the best performance.}
\label{tab: Suppl_N_to_K_ablation_component}
\resizebox{0.7\textwidth}{!}{%
\begin{tabular}{@{}c|cccc|cccc@{}}
\toprule
 & \textbf{Source} & \textbf{PDDA} & \textbf{MTL} & \textbf{TTA} & \textbf{Car} & \textbf{Pedestrian} & \textbf{Cyclist} & \textbf{mAP} \\ \midrule
(a) & \checkmark &  &  &  & 62.69/19.02 & 22.72/18.37 & 20.61/18.13 & 35.34/18.5 \\
(b) & \checkmark & \checkmark &  &  & 72.69/31.83 & 27.1/20.93 & 23.41/19.48 & 41.07/24.08 \\
(c) & \checkmark & \checkmark & \checkmark &  & 71.49/30.18 & 28.53/22.61 & \textbf{24.03}/\textbf{19.57} & 41.35/24.12 \\ \midrule
(d) & \checkmark & \checkmark & \checkmark & \checkmark & \textbf{73.58}/\textbf{33.11} & \textbf{30.01}/\textbf{23.73} & 22.93/18.62 & \textbf{42.17}/\textbf{25.15} \\ \bottomrule
\end{tabular}%
}
\end{table*}

\begin{table*}[t]
\centering
\caption{Component ablations in AP(\%) on Waymo $\rightarrow$ NuScenes. \textbf{Source} represents the conventional training procedure. \textbf{PDDA}, \textbf{MTL} and \textbf{TTA} represent our data augmentation, multi-task learning, and test-time adaptation, respectively. The bold \textbf{values} represent the best performance.}
\label{tab: Suppl_W_to_N_ablation_component}
\resizebox{0.7\textwidth}{!}{%
\begin{tabular}{@{}c|cccc|cccc@{}}
\toprule
 & \textbf{Source} & \textbf{PDDA} & \textbf{MTL} & \textbf{TTA} & \textbf{Car} & \textbf{Pedestrian} & \textbf{Cyclist} & \textbf{mAP} \\ \midrule
(a) & \checkmark &  &  &  & 31.2/19.13 & 10.52/8.39 & 0.75/0.55 & 14.16/9.36 \\
(b) & \checkmark & \checkmark &  &  & 35.67/22.25 & 13.79/10.08 & 1.03/0.79 & 16.83/11.04 \\
(c) & \checkmark & \checkmark & \checkmark &  & 35.69/22.15 & 14.46/10.46 & 1.12/0.89 & 17.09/11.17 \\ \midrule
(d) & \checkmark & \checkmark & \checkmark & \checkmark & \textbf{36.04}/\textbf{22.25} & \textbf{14.48}/\textbf{10.56} & \textbf{1.15/0.95} & \textbf{17.22/11.26} \\ \bottomrule
\end{tabular}%
}
\end{table*}

Table~\ref{tab: Suppl_N_to_K_ablation_component} shows the effect of individual components on detecting all object classes on NuScenes $\rightarrow$ KITTI. It illustrates the significant improvements of the PDDA augmentation on the detection of all object classes and the further improvement of the test-time adaptation on the detection of ``Car'' and ``Pedestrian''.

Table~\ref{tab: Suppl_W_to_N_ablation_component} shows the effect of individual components on detecting all object classes on Waymo $\rightarrow$ NuScenes. It illustrates the significant performance improvements of the PDDA augmentation and the further performance improvement of the proposed test-time adaptation on detecting all object classes.

\end{document}